\DocumentMetadata{uncompress} 
\documentclass[
  10pt,
  longbibliography,
  twocolumn,
  twoside
]{article}

\usepackage[numbers,square,comma,sort&compress]{natbib}
\usepackage[
  left=0.65in,
  right=0.65in,
  top=0.65in,
  bottom=0.65in,
]{geometry}

\usepackage{sectsty}
\sectionfont{\raggedright\large\bfseries}
\subsectionfont{\raggedright\large}

\usepackage[font=small,labelfont=bf]{caption}

\usepackage{fancyhdr}
\usepackage{amssymb}
\usepackage{amsmath}
\usepackage{newunicodechar}
\usepackage{amsfonts}  
\usepackage{array}
\usepackage{nameref}    
\usepackage{hyperref}   
\usepackage{placeins}

\usepackage{environ}

\usepackage{url}
\PassOptionsToPackage{hyphens}{url} 
\makeatletter
\g@addto@macro{\UrlBreaks}{\UrlOrds}
\makeatother

\usepackage[table]{xcolor}   
\usepackage{colortbl}        

\definecolor{goodblue}{RGB}{0, 91, 187}
\hypersetup{
  colorlinks=true,
  allcolors=goodblue,
  urlcolor=goodblue,
  citecolor=goodblue,
  pdfborder={0 0 0},
  breaklinks=true,
}

\usepackage[normalem]{ulem}

\usepackage{textcomp}

\usepackage[export]{adjustbox}

\makeatletter

\def\CT@@do@color{%
  \global\let\CT@do@color\relax
  \@tempdima\wd\z@
  \advance\@tempdima\@tempdimb
  \advance\@tempdima\@tempdimc
  \advance\@tempdimb\tabcolsep
  \advance\@tempdimc\tabcolsep
  \advance\@tempdima2\tabcolsep
  \kern-\@tempdimb
  \leaders\vrule
  \hskip\@tempdima\@plus  1fill
  \kern-\@tempdimc
  \hskip-\wd\z@ \@plus -1fill }
\makeatother

\newcommand{\done}[1]{}

\usepackage{titlesec}
\titleformat*{\paragraph}{\bfseries}

\titlespacing*{\section}{0pt}{3ex}{0.3ex}
\titlespacing*{\subsection}{0pt}{1.5ex}{0.3ex}
\titlespacing*{\subsubsection}{0pt}{1ex}{0.3ex}

\usepackage{changepage}

\usepackage{graphicx}
\usepackage{epsfig}
\usepackage{verbatim}
\usepackage{enumerate}
\usepackage{enumitem}

\usepackage{ifthen}

\usepackage{longtable}

\usepackage{mathtools}

\usepackage{tabularx}

\newboolean{twocolswitch}

\newcommand{\PreserveBackslash}[1]{\let\temp=\\#1\let\\=\temp}

\newcommand{\sindex}[1]{}
\newcommand{\nindex}[1]{}

\newcommand{\www}[1]{\url{#1}}

\usepackage{lettrine}

\usepackage{changepage}

\usepackage{footmisc}

\usepackage{refcount}

\makeatletter
\newcommand{\footnotemarklabel}[1]{%
  \protect\footnotemark
  \begingroup
    \edef\@currentlabel{\thefootnote}
    \label{#1}%
  \endgroup
}
\makeatother

\renewcommand{\footnoterule}{%
  \kern 3pt
  \hrule width 0.4\columnwidth height 0.4pt
  \kern 6pt
}
\NewEnviron{excerpt}{
    \begin{quote}
      \medskip
      \BODY
      \medskip
    \end{quote}
}

\NewEnviron{editnote}{
    \begin{quote}
      \color{rose}
      $\blacksquare$
      \BODY
      \medskip
    \end{quote}
}

\newcommand{\command}[1]{
  \lstinline[language={[LaTeX]TeX},basicstyle=\ttfamily]{#1}
}

\newcommand{\editbox}[2]{
}

\newcommand{\editboxwithlatex}[2]{
}

\usepackage{fancyvrb}

\usepackage{tcolorbox}
\tcbuselibrary{skins,breakable,listings,breakable}

\usepackage{tikz}
\usetikzlibrary{shapes}

\tikzstyle{mybox} = [draw=lightblue!70, fill=lightblue!7, very thick,
    rectangle, rounded corners, inner sep=10pt, inner ysep=20pt]

\tikzstyle{editortitle} =[draw=archetyperowcoloralt, fill=archetyperowcoloralt, text=black]

\newcommand\Loadedframemethod{default}
\usepackage[framemethod=\Loadedframemethod]{mdframed}

\mdfsetup{skipabove=\topskip,skipbelow=\topskip}

\tikzstyle{loglinetitle} =[draw=icedark, fill=icemedium!50, text=black]

\newenvironment{loglinebox}[1][]{

  \ifstrempty{#1}%
  {\mdfsetup{%
    frametitle={%
       \tikz[baseline=(current bounding box.east),outer sep=0pt]
        \node[loglinetitle, anchor=east,rectangle]
        {\strut~~#1~~\strut};}}
  }%
  {\mdfsetup{%
     frametitle={%
       \tikz[baseline=(current bounding box.east),outer sep=0pt]
        \node[loglinetitle,anchor=east,rectangle]
        {\strut~~#1~~\strut};}}%
   }%
   \mdfsetup{innertopmargin=5pt,linecolor=icedark,%
             linewidth=0.5pt,topline=true,
             frametitleaboveskip=\dimexpr-\ht\strutbox\relax,}
   \begin{mdframed}[backgroundcolor=icelight,nobreak=true]\relax%
     \raggedright
}{\end{mdframed}}

\tikzstyle{abstracttitle} =[draw=magmadark!75, fill=magmamedium!75, text=black]

\newenvironment{abstractbox}[1][]{

  \ifstrempty{#1}%
  {\mdfsetup{%
    frametitle={%
       \tikz[baseline=(current bounding box.east),outer sep=0pt]
        \node[abstracttitle, anchor=east,rectangle]
        {\strut~~#1~~\strut};}}
  }%
  {\mdfsetup{%
     frametitle={%
       \tikz[baseline=(current bounding box.east),outer sep=0pt]
        \node[abstracttitle,anchor=east,rectangle]
        {\strut~~#1~~\strut};}}%
   }%
   \mdfsetup{innertopmargin=5pt,linecolor=magmadark,%
             linewidth=0.5pt,topline=true,
             frametitleaboveskip=\dimexpr-\ht\strutbox\relax,}
   \begin{mdframed}[backgroundcolor=magmalight,nobreak=true]\relax%
     \raggedright
}{\end{mdframed}}

\tikzstyle{infotitle} =[draw=darkgrey, fill=lightgrey!50, text=black]

\newenvironment{infobox}[1][]{

  \ifstrempty{#1}%
  {\mdfsetup{%
    frametitle={%
       \tikz[baseline=(current bounding box.east),outer sep=0pt]
        \node[infotitle, anchor=east,rectangle]
        {\strut~~#1~~\strut};}}
  }%
  {\mdfsetup{%
     frametitle={%
       \tikz[baseline=(current bounding box.east),outer sep=0pt]
        \node[infotitle,anchor=east,rectangle]
        {\strut~~#1~~\strut};}}%
   }%
   \mdfsetup{innertopmargin=5pt,linecolor=grey,%
             linewidth=0.5pt,topline=true,
             frametitleaboveskip=\dimexpr-\ht\strutbox\relax,}
   \begin{mdframed}[backgroundcolor=lightgrey!25,nobreak=true]\relax%
     \raggedright
}{\end{mdframed}}

\tikzstyle{essencetitle} = [draw=magmadark!75, fill=magmamedium!75, text=black]

\usepackage{enumitem}

\tikzstyle{changelogtitle} =[draw=darkgrey, fill=lightgrey!50, text=black]

\usepackage[table]{xcolor}

\definecolor{olivegreen}{rgb}{0.33333,.41961,0.18431}
\definecolor{forestgreen}{rgb}{0.13333,.5451,0.13333}

\definecolor{lightgrey}{rgb}{0.7,0.7,0.7}
\definecolor{verylightgrey}{rgb}{0.90,0.90,0.90}
\definecolor{veryverylightgrey}{rgb}{0.95,0.95,0.95}
\definecolor{grey}{rgb}{0.5,0.5,0.5}
\definecolor{darkgrey}{rgb}{0.3,0.3,0.3}
\definecolor{verydarkgrey}{rgb}{0.15,0.15,0.15}

\definecolor{headerblue}{HTML}{33367E}
\definecolor{unitednationsblue}{HTML}{4D88FF}

\definecolor{charcoal}{HTML}{36454F}
\definecolor{cinerous}{HTML}{98817B}
\definecolor{feldgrau}{HTML}{4D5D53}
\definecolor{glaucous}{HTML}{6082B6}
\definecolor{arsenic}{HTML}{3B444B}
\definecolor{xanadu}{HTML}{738678}

\definecolor{firebrick}{HTML}{B22222}
\definecolor{orangered}{HTML}{FF4500}
\definecolor{tomato}{HTML}{FF6347}

\definecolor{orange}{RGB}{255,116,0}

\definecolor{purpletaupe}{HTML}{3B444B}

\definecolor{rose}{HTML}{E3242B}

\colorlet{editnotecolor}{rose}

\definecolor{headerorange}{RGB}{255,116,0}
\definecolor{headergray}{RGB}{230,230,230}

\definecolor{headerpop}{RGB}{230,230,230}

\definecolor{magmalight}{RGB}{252,251,195}
\definecolor{magmalightalt}{RGB}{250,240,184}
\definecolor{magmamedium}{RGB}{245,200,146}
\definecolor{magmadark}{RGB}{224,106,98}

\definecolor{icelight}{RGB}{223,242,244}
\definecolor{icelightalt}{RGB}{189,222,226}
\definecolor{icemedium}{RGB}{132,184,204}
\definecolor{icedark}{RGB}{103,153,191}

\definecolor{datasetrowcolor}{RGB}{232,244,234}
\definecolor{datasetrowcoloralt}{RGB}{210,231,214}

\newcommand{\heapslaw}{Heaps' law}

\newcommand{\candor}{CANDOR}
\newcommand{\movies}{Movies (individual)}
\newcommand{\scripts}{Movies (grouped)}
\newcommand{\moviesindivshort}{Movies (I)}
\newcommand{\moviesgroupshort}{Movies (G)}

\newcommand{\convoNlist}{\candor~$=1,455$, \movies~$=3,104$, \scripts~$=589$}

\newcommand{\heapsexponent}{\beta}

\newcommand{\standrderrorstatement}{Standard error of the slope: no notation for SES $<.01$, \string^ for $.05 > \text{SES} \geq .01$, and $\ast$ for SES $> .05$}

\setboolean{twocolswitch}{true}

\raggedright

\usepackage{pdfpages}
\usepackage{newpax}
\newpaxsetup{usefileattributes=true}

\usepackage{authblk}

\begin{document}

\title{\protect
Statistical laws and linguistics differ in
\\
naturalistic video and fictional conversations








}


\renewcommand*{\Authsep}{, }
\renewcommand*{\Authand}{, }
\renewcommand*{\Authands}{, }
\renewcommand*{\Affilfont}{\normalsize\normalfont}
\renewcommand*{\Authfont}{\bfseries}
\setlength{\affilsep}{2em}

\author[1,2\thanks{ashley.fehr@uvm.edu}]{Ashley M. A. Fehr}
\author[1,2]{Calla G. Beauregard}
\author[1,2,3]{Julia Witte Zimmerman}
\author[4]{Katie Ekstr{\"o}m}
\author[1,5]{Pablo {Rosillo-Rodes}}
\author[1,2,6]{Christopher M. Danforth}
\author[1,2,7,8\thanks{peter.dodds@uvm.edu}]{Peter Sheridan Dodds}

\affil[1]{
  Computational Story Lab,
  Vermont Advanced Computing Center,
  University of Vermont,
  Burlington,
  VT 05405,
  US
}

\affil[2]{
  Vermont Complex Systems Institute,
  MassMutual Center of Excellence for Complex Systems and Data Science,
  University of Vermont,
  Burlington,
  VT 05405,
  US
}

\affil[3]{
  Computational Ethics Lab,
  University of Vermont,
  Burlington,
  VT 05405,
  US
}

\affil[4]{
  Glass Brain Lab, Department of Biomedical Engineering; Vermont Conversation Lab, Larner College of Medicine,
  University of Vermont,
  Burlington,
  VT 05405,
  US
}

\affil[5]{
  Institute for Cross-Disciplinary Physics and Complex Systems IFISC (UIB-CSIC),
  Campus Universitat de les Illes Balears,
  E-07122 Palma de Mallorca,
  Spain
}

\affil[6]{
  Department of Mathematics \& Statistics,
  University of Vermont,
  Burlington,
  VT 05405,
  US
  }

\affil[7]{
  Santa Fe Institute,
  1399 Hyde Park Rd,
  Santa Fe,
  NM 87501,
  US
}

\affil[8]{
  Department of Computer Science,
  University of Vermont,
  Burlington,
  VT 05405,
  US
}

\date{\today}

\maketitle


\hspace*{30pt}
\begin{minipage}{600pt}

  \begin{tabular}{>{\raggedright\arraybackslash}m{275pt}p{15pt}m{150pt}}

    \parbox{275pt}{
      \begin{abstractbox}[Abstract]
        \raggedright
        Conversation is a cornerstone of social connection and is linked to well-being outcomes. Conversations vary widely in type with some portion generating complex, dynamic stories. One approach to studying how conversations unfold in time is through statistical patterns such as Heaps' law, which holds that vocabulary size scales with document length. Little work on Heaps' law has looked at conversation and considered how language features impact scaling. We measure Heaps' law for conversations recorded in two distinct mediums:
1. Strangers brought together on video chat
and
2. Fictional characters in movies.
We find that scaling of vocabulary size differs by parts of speech, suggesting a less efficient purpose in communication by medium.
        \smallskip
      \end{abstractbox}
    }

    \parbox{275pt}{
      \begin{infobox}[Keywords]
        \raggedright
        conversation,
naturalistic,
dialogue,
video chat,
movies,
Heaps' law,
Zipf's law,
parts of speech,
linguistics

        \smallskip
      \end{infobox}
    }
    &

    &
    \parbox{150pt}{
      \begin{loglinebox}[Logline]
        \raggedright
Conversations appear different by medium. Statistical laws help show these differences by parts of speech in naturalistic and fictional conversations. 
        \smallskip
      \end{loglinebox}
    }
    \vspace{2mm}

    \parbox{150pt}{
      \includegraphics[width=\linewidth,valign=m,frame]{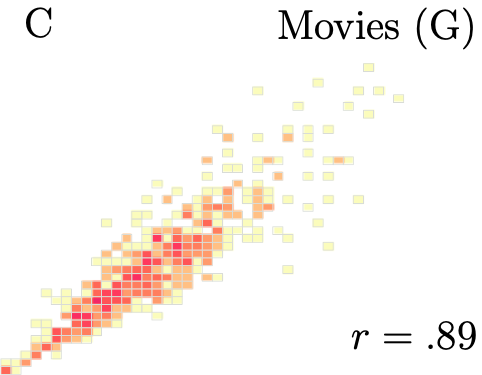}
    }

  \end{tabular}

\end{minipage}


\newgeometry{
  left=2in,
  right=2in,
  top=1in,
  bottom=1in,
  }

\onecolumn

\renewcommand{\baselinestretch}{1.25}
\selectfont

\renewcommand{\baselinestretch}{1}
\selectfont

\twocolumn

\restoregeometry


\clearpage



\section{Introduction}
\label{sec:papertag.introduction}

Language is social, conveying meaning and understanding between individuals and groups~\cite{clark1996using_lang}.
We employ language in conversation to fulfill both inter- and intra-personal needs, supporting communication, collaboration, and social connection~\cite{baumeister1995needtobelong, benus2014social_entrainment}. In addition, \textit{how} people engage in conversation can be linked to public health outcomes such as well-being~\cite{sun2020_wellbeing, wheatley2024}, motivating the study of conversation at scale~\cite{holt2017soc_connection_priority}.
Conversation participants mutually determine the trajectory of their exchange, weaving a complex dynamic story~\cite{soloman2023, carragher2024coconstructedcomm,clark1996using_lang} in which singular processes may be insufficient to describe outcomes ~\cite{paletz2023speaking_similarly}. Therefore, methods from the analysis of dynamical systems can help capture emergent properties that remain elusive at the individual scale.

Linguistic scaling offers a way to understand more about the shape of information and symbols that underlie the construction of meaning in conversation.
With linguistic scaling, we can explore how statistical patterns can quantify and ultimately illuminate social processes.
Previous work on statistical scaling in text (e.g., Heaps' law and Zipf's law) is dominated by long-form, monologic text such as the Google Books corpus~\cite{petersen2012langs_cool_as_they_expand,pechenick2017lexicalturbulencescaling} and literary works~\cite{chacoma_tags_2020,gerlach_vocabgrowth2013,bian2016scaling_like_lin,altmann2024statisticallaws,williams2015text_mixing_rank_freq,williams2016zipf_coherent_lang_production,williams2015zipfs_phrases}, leaving open questions as to how prior observations extend to conversational mediums in particular.

We study conversational datasets, comparing naturalistic video conversation to fictional conversation (dialogue exchanges in movie scripts) to explore how linguistic units differ.
The specific gaps we hope to address include the applications of analytic methods from dynamical systems, particularly through the lens of \heapslaw\ and Zipf's law, with attention to potential mechanisms for differences by conversational medium.

\subsection{Conversation and medium}
\label{sec:papertag.introduction.convo_and_medium}

Conversation is co-constructed~\cite{carragher2024coconstructedcomm,ruhlemann2007ConversationContext} dialogue between two or more participants~\cite{yeomans2023guide_to_convo_research, yeomans2022convo_circumplex, brooks2025talk} where turn-taking~\cite{sacks1974turntaking_fundamental, stivers2009turntaking_universal} and its coordination~\cite{meyer2023timing_natural_convo,ruhlemann2007ConversationContext} are fundamental features.
Broadly, we construe conversation to include forms of dialogue or discourse that vary beyond the face-to-face, spoken modality. Thus, conversation substrates include spoken or naturalistic conversation, texting or direct messages, fictional or scripted conversation, and discussion forums.
Conversation is also affected by the modality~\cite{boudin2024multimodal_feedback_in_convo} and the technology through that transmits it~\cite{williams1974tv_technology_culture,mcluhan1964medium}, so our broad definition facilitates exploration across mediums
where the medium may facilitate differences in conversation structure and function.

Conversation medium affects communication outcomes because of unique limitations and affordances of each substrate.
Many mediums act as conversation substrates, for example, by defining the limitations on how much communication can occur (character limits or post frequency), which topics (moderation), what editing is enabled, and who the participants are. In defining these communication aspects, the medium shapes the structure of the conversation.

In this work, we compare naturalistic (from video) to scripted conversation because of various properties that may give rise to structural differences.
Naturalistic or spoken conversation has vocal and gestural features and is spontaneously structured while scripted conversation includes dialogue that is edited or cognitively prepared by a third party. When transcribed, spoken conversation data is flattened in contrast to scripts where it is instantiated in text and acted. 
In both cases, some meaningful signal is not present in the text, such as prosodic contours which carry pragmatic information that is only minimally presented in text through the use of punctuation.
Although scripted conversation's written form might suggest it is more akin to other written texts, such as novels, it has some similarities with naturalistic conversation. Importantly, both substrates center the resolution of a speaker's intended meaning and audience understanding~\cite{clark1996using_lang}, an important role in narrative and storytelling.
These considerations motivate our interest in these mediums to determine how analysis of transcribed text may reveal structure. Next we describe how statistical scaling has been applied in the study of texts.

\subsection{Scaling in text}
\label{sec:papertag.introduction.scaling}

For text corpora,
\heapslaw\
classically describes a scaling relationship between the total number of words---tokens---and the number of unique words---types~\cite{heaps1978InformationRetrieval,rosillo_rodes2025heap_paper}.
For example, when reading through the first $t$ words of a book,
the number of unique words $N(t)$
scales as
$N(t) \sim t^{\beta}$
where $
0 \le \beta \le 1$.
Types collectively comprise a text's vocabulary $N(t)$.
A vocabulary may also contain non-words such as numbers and punctuation.
More generally for complex systems, \heapslaw\ is an instance of a type-entity relationship
(e.g., tree species and individual trees).

\heapslaw\ is understood to arise for growing systems which comprise many types that follow Zipf's law~\cite{rosillo_rodes2025heap_paper}.
For language, Zipf's law is a power-law frequency-rank relationship expressed as
$S_{r} \sim r^{-\alpha}$ where
$r$ is the rank of any given word and
$S_{r}$ is its count (and for general systems, $S$ indicates size).
While the connection between Zipf's law and Heaps' law
is not a simple scaling for finite systems,
in the limit of large texts~\cite{rosillo_rodes2025heap_paper},
$\beta = 1/\alpha$
~\cite{rosillo_rodes2025heap_paper,baeza-yates_block_1997}.

We focus our findings around Zipf's law and \heapslaw, as word counts are commonly used in natural language processing (NLP).
We make the assumption that text follows the form $N(t) \sim t^{\beta}$, evidence of which we describe next. While goodness of fit of this model and the underlying power-law distribution are often the subject of statistical scaling law research, our focus is on linguistic discussion; thus, we de-focus on goodness of fit.


\heapslaw\ research has focused on written language heavily based in literary works or corpora~\cite{chacoma_tags_2020, bian2016scaling_like_lin,altmann2024statisticallaws}, such as Google's N-gram books corpus~\cite{gerlach_vocabgrowth2013}.
Some literary work examined scaling by parts of speech (nouns and verbs)~\cite{chacoma_tags_2020,piantadosi2014zipf_mechs_review} and lemmas (root word forms)~\cite{corral2015lemmas_zipf}, confirming good fit of both linguistic units as is often found with word unigrams.
Outside of classical literature, some textual analysis has looked at Wikipedia pages~\cite{altmann2024statisticallaws}, internet forums~\cite{kubo_2011}, news across languages~\cite{yu2018zipf_50langs_structural_pattern,rosillo_rodes2025entropy_ttr_gigaword}, and Twitter~\cite{rosillo_rodes2025entropy_ttr_gigaword} to replicate Zipf's or \heapslaw.

Relatively less text scaling research has focused on conversational corpora, and even fewer works have offered social linguistic insight into scaling mechanisms.
A notable work in this niche studied multilingual naturalistic corpora in mixed settings (face-to-face, telephone, and task-based), looking at
parts of speech and open and closed classes~\cite{linders2023zipf_spoken_dialog}. Ref.~\cite{linders2023zipf_spoken_dialog} described the scaling mechanism in spoken dialogue as stemming from the principle of least effort. Other relevant studies have focused on spoken language transcripts such as elderly group conversation~\cite{abe_2021} and
a small transcribed naturalistic sample from the British National Corpus in comparison to books~\cite{bian2016scaling_like_lin}.

The monologic counterpart has usually been literary works focusing on word unigrams~\cite{linders2023zipf_spoken_dialog}, with little research beyond words~\cite{chacoma_tags_2020,linders2023zipf_spoken_dialog,yu2018zipf_50langs_structural_pattern}. It is worth exploring naturalistic conversation with respect to a non-naturalistic conversational counterpart because we would expect structural differences beyond contrasts with literary works.
These differences could help us further understand the uniqueness of naturalistic conversation---what makes spoken conversation naturalistic and other forms less so?
Thus, we see our work filling a gap in qualifying why there may be differences in conversational medium.
One mechanism that may help understand differences by medium is linguistic efficiency.


\subsection{Linguistic theory and tools}
\label{sec:papertag.introduction.linguistic_theory}


Numerous domains interrogate how efficiency shapes language.
Zipf described the speaker preference to communicate efficiently as the principle of least effort~\cite{zipf1949least_effort}.
Work in information efficiency and psycholinguistics suggests it shapes the structure of language~\cite{futrell2024linguistic_info_efficiency,gibson2019efficiency_shapes_lang,zimmerman_tokens_2024, petersen2012langs_cool_as_they_expand} where human preference for communicative efficiency shapes language choices~\cite{fedzechkina2012restructure_lang_efficiency,jaeger2011efficiency_shape_lang_product}.
And in pragmatic linguistics, this efficiency can be seen in Grice's maxim of quantity in conversation in which one communicates succinctly~\cite{grice1989wayofwords}, a heuristic which is supportive of linguistic scaling~\cite{xu2024wordreuse}.


From such works, we derive that (1) one significant purpose of a lexicon is to encode novel ideas---people do this through re-use and re-combination of words~\cite{xu2024wordreuse} and (2) efficiency, sometimes invoked as the principle of least effort~\cite{zipf1949least_effort} shapes the structure of communication~\cite{ferrer_i_cancho2003least_effort}.
In particular, the maxim of quantity suggests how speakers tie language to efficiency---they balance pressures to both minimize cognitive load and maximize communicative diversity to achieve listener comprehension~\cite{grice1989wayofwords,futrell2024linguistic_info_efficiency}.
Conversation is an interesting setting for achieving this balance in which we can apply NLP tools to measure potential mechanisms. 

Next, we describe multiple linguistic units of meaning in conversation that we use to explain scaling.
In addition to unigrams, we investigate the scaling of parts of speech (POS) to better understand their roles in meaning and scaling production by medium. Parts of speech serve different communicative purposes in any text. Nouns and verbs primarily convey content whether new or referential to content already mentioned. Function words provide grammatical structure and connection in language, signaling meaning through interactions with other words. 
Interjections are exclamatory and not as syntactically related to surrounding parts of speech~\cite{universaldependencies2021}.
Parts of speech belong to either an `open' or a `closed' word class. The open class contains content words and new nouns and verbs can be added. The closed class is more fixed, containing a relatively static inventory of function words and numerals~\cite{universaldependencies2021}. We expect open class POS to show more novel types and therefore higher scaling exponents than closed class POS.

We examine interjections more closely for their importance in conversations.
Interjections are a social linguistic gesture in the form of word (e.g., \textit{wow}, \textit{really}, \textit{oh my god}), a non-word (e.g., \textit{pffft}), or a phrase that can be indicative of a mental state or an emotive attitude~\cite{ameka1992interjections}, reinforcing moments in a story or mirroring a speaker's emotion~\cite{cooney2024naturalturn}. In this way, they are a significant source of interlocutor feedback~\cite{boudin2024multimodal_feedback_in_convo}, even in text form which may have some gestural meaning encoded. That is, spoken interjections may pair with physical gestures that we have come to intuit as emotive feedback when encountering specific interjections.
We do not expect interjections to appear frequently outside of transcribed, spontaneous conversation
given both interjections' described roles and the constraints of scripted authorship.

\section{Current work}
\label{sec:papertag.current_work}

We explore \heapslaw\ in naturalistic video conversations between strangers as well as in fictional movie conversations, building on prior work by comparing across mediums and offering descriptive and exploratory insights to scaling differences.
We utilize statistics such as POS proportions and the type-token ratio, a metric in lexical diversity that has been used to compare languages~\cite{gutierrez_vasques2019morph_complexity,rosillo_rodes2025entropy_ttr_gigaword} and is directly related to \heapslaw\ as diversity drives vocabulary size.
Last, we explore temporal word distributions~\cite{goh2008burstiness_memory,altmann2009beyond_word_freq} to understand how interarrival times of types and POS may correspond in scaling and proportion results. 
Previous work has shown that naturalistic verbal communication exhibits irregular and clustered arrivals
~\cite{abney2018verbal_nonverbal_bursts}. 
Analyzing temporal and scaling behavior together may therefore have implications for the two mediums in our investigation which we have not seen.

Broadly, we expect differences between the naturalistic and non-naturalistic corpora through \heapslaw\ and descriptive statistics, with exploratory analysis qualifying scaling differences between conversation mediums.
We expect parts of speech scaling and usage to differ between open and closed word classes, and we expect interjections to appear more frequently in naturalistic (video) conversation.

\section{Methodology}
\label{sec:papertag.methods}

\subsection{Data sets}
\label{sec:papertag.data}

We use the Conversation: A Naturalistic Dataset of Online Recordings (CANDOR) dataset~\cite{reece2023candor} and the Movie-Dialogs Corpus~\cite{danescu-niculescu-mizil2011ChameleonsImaginedConversations}.
We reference two formulations of data from the Movie-Dialogs Corpus: 
``\movies'' as meaning individual conversations among speakers whereas 
``\scripts'' means all conversations in an entire movie have been joined;
we refer to these corpora as \moviesindivshort\ and \moviesgroupshort\ for brevity. 
We provide dataset description and specific cleaning and subsetting methods in each corpus subsection. We distinguish between `word' and `type' when counts refer specifically to words versus inclusive of all tokenized `types' such as punctuation.

\subsubsection{\candor\ corpus}

\candor\ data were collected in 2020~\cite{reece2023candor}. The corpus contains $1,456$ unique participants. All participants were adults of ages 18 to 66, and some participants completed more than one conversation.
We remove one conversation as
it was shortened because of technical difficulties. The final count of unique participants is $1,455$.
The conversation data contains $1,655$ English language conversations, $315,938$ conversation turns (utterances), and $9.2$M word tokens.

The conversation data exists pre-parsed into utterances using four different turn segmentation models: Audiophile, Cliffhanger, Backbiter~\cite{reece2023candor}, and NaturalTurn~\cite{cooney2024naturalturn}. The choice of turn segmentation model will change the number of utterances, pauses, gaps, and counts of other between-turn variables. This project uses the NaturalTurn model which joins a speaker's intended utterances while preserving a listener's parallel speech as separate utterances, resulting in fewer utterances than prior segmentation models applied to \candor.

\subsubsection{Movie-Dialogs Corpus: \moviesindivshort\ and \moviesgroupshort}

The Movie-Dialogs Corpus~\cite{danescu-niculescu-mizil2011ChameleonsImaginedConversations} contains fictional conversations diarized into ordered utterances from movie scripts. Release years range from 1927 to 2010 with a heavy skew to the 2000s.
Movies include $166$ conversations with: less than $10$ utterances, $2$ non-fiction genres, and $274$ null utterances fields. We remove conversations having these qualities for more robust signal and analysis.
The final corpus contains $2,397$ unique characters from $472$ unique movies throughout the release years 1927 to 2010. The conversation data include $3,104$ conversations, $43,762$ utterances, and $485,389$ words. This corpus forms our \movies\ comparison group.

Movie conversations are very short compared to CANDOR's naturalistic ones, so we formulate another exploration of Movie scaling to look at all conversations within one movie as a `conversation.'
The conjoined speaking parts formulation as another `conversation' conveys plot solely through dialogue from numerous participants. Still excluding conversations with null utterances or movies in non-fiction genres, this formulation includes $8,481$ unique characters, $589$ movies (as $589$ conversations), totaling $288,125$ utterances and $3$M words. We refer to this comparison group as \scripts.

\subsection{Tools}
\label{sec:papertag.tools}

All coding occurs in a Python 3.12 environment in Visual Studio Code, with libraries including NumPy, Pandas, and Matplotlib.
Statistical analyses rely on SciPy for linear regression. NLP relies on regular expressions (\texttt{re} library) and NLTK for tokenization while Stanza's pretrained English model~\cite{qi_stanza_2020} provides part of speech (POS) classification. Stanza's model was trained on treebanks (containing linguistic dependency information), and its POS processor uses sentence context in POS tag classification.

\subsection{Procedure}

\subsubsection{Data cleaning}
We requested and received \candor\ data from its respective authors~\cite{candor_data_url} and downloaded the Movie-Dialogs dataset in its legacy format~\cite{movie_dialogs_data_legacy_url}.

Data cleaning includes various NLP steps depending on the conversation mode to bring the written artifacts in movie scripts more in line with spoken conversation in \candor. 
For \candor, we remove a transcription artifact in NaturalTurn (`$\ll$' and `$\gg$'). 
For movies, 
we standardize quotes and hyphens and fix some encoding issues and whitespace issues. For all data, we spell out shortened word-forms: sorta, dunno, gonna, wanna, gotta. The last NLP step is Stanza's POS tagging, with recoding tags to fewer categories (see Appendix~\ref{sec:appendix.methods_extra}). For one of our exploratory variables, we code conversations as having low or high interjection usage using a median cut within corpus.

After careful consideration of punctuation's place in the text form of conversation,
we retain punctuation. Punctuation can convey meaning and has statistical importance in showing Zipf's law~\cite{stanisz2024nlp_complexity}. Removing punctuation mildly influences POS tagging and subtly changes scaling coefficients while increasing standard error of the slopes. For instance, interjection (INTJ) occurrences are 1.5M without punctuation and shrink to 598k with punctuation because punctuation acts as context during classification.
It is possible that punctuation removes INTJs that are spoken as utterances (e.g., ``ok'') and could serve a continuer purpose.

\subsubsection{Statistical procedures}
We use Pearson correlations, observe distributions, and analyze descriptive statistics.
We also explore temporal measures of words and POS tags with token interarrival times, burstiness, and memory~\cite{altmann2009beyond_word_freq,goh2008burstiness_memory}. Interarrival times are gaps between successive occurrences of types. Burstiness codifies the variation of interarrival times in a metric ($-1<\text{B}<1$), with $\text{B}>0$ suggesting `bursty' behavior (less regular) while $\text{B}<0$ indicates more regular behavior. Memory is a correlation measure of interarrival times that codifies if these gaps are independent or show memory ($-1<\text{M}<1$). $\text{M}>0$ indicates that similar length gaps follow each other (short and short or long and long). $\text{M}<0$ suggests more alternation between short and long gaps.
Last, we calculate the type-token ratio (TTR) as a measure of lexical diversity or richness~\cite{gutierrez_vasques2019morph_complexity,rosillo_rodes2025entropy_ttr_gigaword}. TTR is defined as vocabulary size (unique types) divided by total words, with a value closer to $1$ reflecting higher diversity.

We assess \heapslaw\ and Zipf's law with least-squares linear regressions, identifying the upper scaling regime for Heaps' scaling behavior. We collect total and unique types over the length of the document. We visually identify a restricted range of the data to isolate the sublinear scaling regime in log-log space. The heavy tail of word frequency distributions
presents some measurement difficulty, one of which is
can violate the normality assumption underlying least squares regression. Restricting the regime allows us to still use regression to study the steepness of slopes~\cite{clauset2009powerlaw_regimes}.

We apply a horizontal axis restriction in log-space for the Zipf regressions (ranking of types) and correspondingly apply those same restrictions to Heaps' regressions' vertical axis (unique types).
These restrictions better align the expected correspondence between Heaps' and Zipfian findings.
Given the shortness of data in \moviesindivshort\ and some POS, we could not apply the same $x_{\text{max}}$ restriction to all analyses (they would be null if there were not sufficient data). We define start- and end-range matrices to specify values per analysis (in Appendix~\ref{sec:appendix.methods_extra}). Further,
the average length of documents makes some plots' lines appear longer or shorter; in particular, \moviesindivshort\ has much shorter documents, leading to their total `types' (plot line) falling short of the other corpora.

Next, we run a least-squares linear regression per corpus, POS by corpus, and low versus high interjection usage by corpus.

\section{Results}
\label{sec:papertag.results}


\subsection{Descriptive analyses}
\label{sec:descriptives}


\begin{table}[b!]
    \centering
    \begin{tabular}{c|c|c|c|}
    \textbf{} & \textbf{\candor} & \textbf{\moviesindivshort} & \textbf{\moviesgroupshort} \\
    \hline
    \textbf{\shortstack{Utterances \\ {}}}
    & \shortstack{$191$ \\ $(66.42)$}
    & \shortstack{$14$ \\ $(5.56)$}
    & \shortstack{$489$ \\ $(221.55)$} \\

    \textbf{\shortstack{Words \\ {}}}
    & \shortstack{$5563$ \\ $(1556.26)$}
    & \shortstack{$156$ \\ $(92.49)$}
    & \shortstack{$5122$ \\ $(2463.41)$} \\

    \textbf{\shortstack{Speaker \\ words}}
    & \shortstack{$6327$ \\ $(5459.16)$}
    & \shortstack{$202$ \\ $(268.94)$}
    & \shortstack{$356$ \\ $(560.50)$} \\

    \end{tabular}
    \hspace{4em}
    \caption{Raw utterances and words by corpus and words by speaker, reported as: $\mu$ ($\sigma$). See Appendix~\ref{sec:appendix.descrip} for word by utterance relationships in conversations and more spread statistics.}
    \label{table:descriptive_stats}
\end{table}

\begin{figure}[t!]
    \centering
    \includegraphics[width=.8\columnwidth]{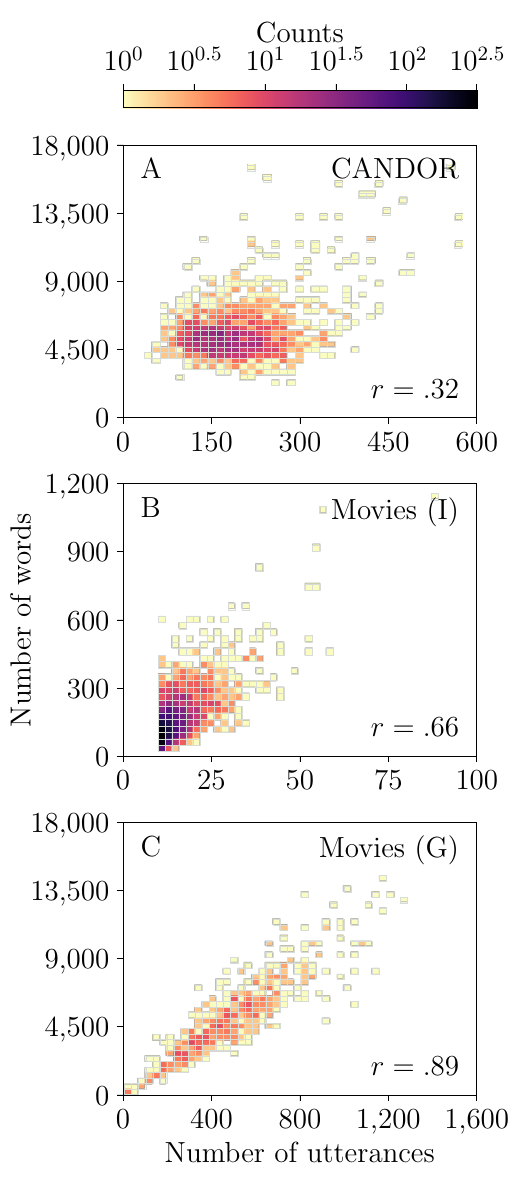}
    \caption{Words by utterances heatmap per corpus with Pearson correlation coefficients. The \movies\ corpus is subsetted to include conversations having a minimum of 10 utterances. Cell count coloring is in log$_{10}$ space, with the global cell max $=311$.}
    \label{fig:all_scatter_aligned}
\end{figure}

Corpora descriptive statistics ($\mu,~\sigma$) for number of words and utterances at the conversation level can be found in Table \ref{table:descriptive_stats}.
\candor\ and \moviesgroupshort\ result in the most similar structure for number of words while \moviesindivshort\ has distinctly fewer words and utterances.
Given the wide-ranging variance, we look at coefficients of variation ($\sigma/\mu$) to quantify the spread.
Of the six categories (words and utterances per corpus), number of words for \moviesindivshort\ ($0.59$) and \moviesgroupshort\ ($0.48$) shows the highest coefficients of variation (compared to $\text{\candor} = 0.28$). 
Related, we average across conversations' type-token ratios (TTRs) by corpus. The corpus TTR distributions are normally distributed, and we report TTR as $\mu~(\sigma)$. Each corpus's ratio appears distinct, with \candor\ ($0.17~(0.02)$) having the lowest diversity, followed by \moviesgroupshort\ ($0.24~(0.07)$) and \moviesindivshort\ ($0.64~(0.09)$).

Further, the relationship between number of words and number of utterances is much stronger in Movies (grouped: $r=0.89$, individual: $r=0.67$) than in \candor\ ($r=0.32$). Movie text appears less consistent and more subjective to outliers than \candor. Overall, \moviesgroupshort\ is distinctly linear (as seen in Figure~\ref{fig:all_scatter_aligned}).

\begin{figure*}[t]
    \centering
    \includegraphics[width=.9\textwidth]{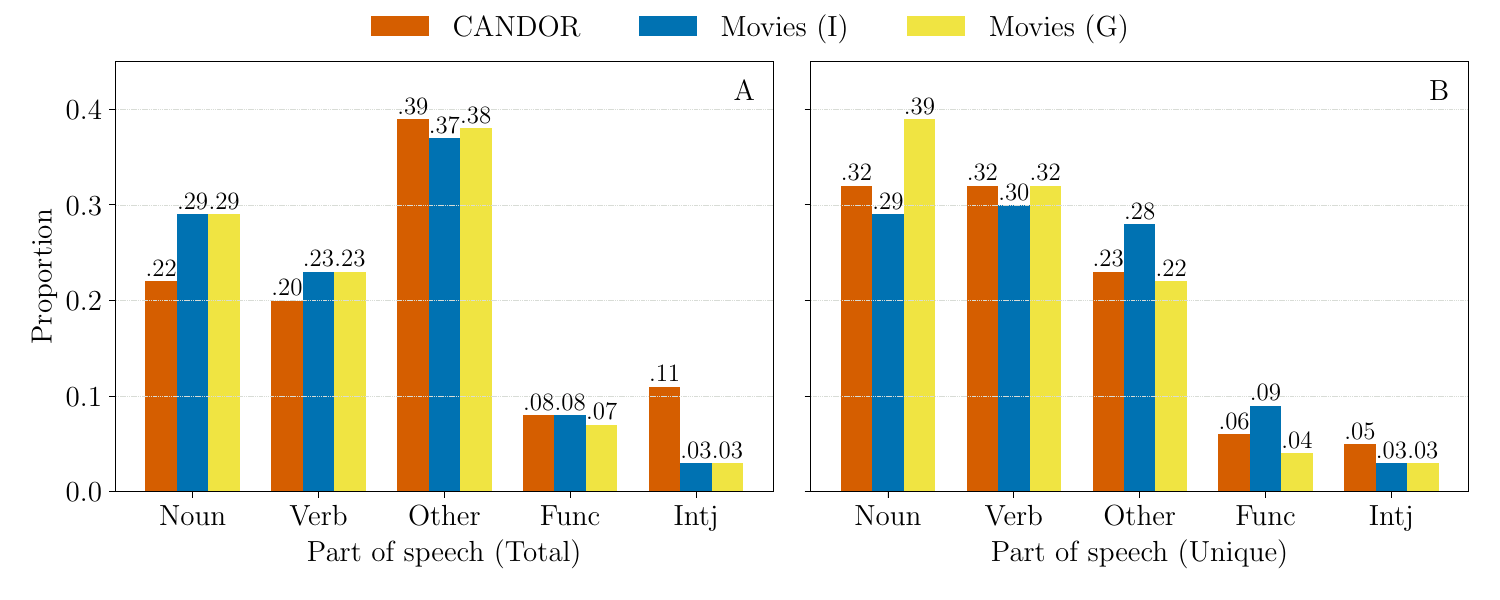}
    \caption{Normalized part of speech corpus proportions as total words (A) and unique words (B). A corpus' total and unique proportions each sum to $1.0$ within corpus.}
    \label{fig:pos_descriptives_normed}
\end{figure*}

Next, we analyze conversations' unique word stagnation to detect outliers. A `run' of unique words indicates how many words before stagnation of new words occurs. We first identify that the distributions of conversations' maximum runs are right (positively) skewed toward shorter runs. \candor\ ($55$ words) and \moviesgroupshort\ ($48$) show very similar run medians with more variance for movie scripts. \moviesindivshort\ ($6$) skews the shortest. We identify the run value at which $99.95\%$ of data is included in the distribution and detect outliers over this cut-off (Figure~\ref{fig:regression_outliers_panel}). \candor's outlier is one of the corpus's longest conversations. \moviesindivshort's outliers include a singing exchange from The Rocky Horror Picture Show and a scene from The Leopard Man; we note that The Leopard Man includes an alternate scene included in this film corpus that partially duplicates its main scene.
 The Rocky Horror Picture Show exchange shows the same lines being exchanged between the character Rocky and guests, resulting in low novelty. \moviesgroupshort's outlier is the dramatic film, Magnolia.

Parts of speech proportions per corpus are in Figure~\ref{fig:pos_descriptives_normed} (with results split by total versus unique words in Table~\ref{table:pos_descriptives}).
Despite \moviesgroupshort\ often having the highest variance across POS, \candor\ and \moviesgroupshort\ appear more similar in some categories than does \moviesgroupshort\ to \moviesindivshort.

\begin{figure*}[t]
    \centering
    \begin{minipage}[c]{0.38\textwidth}
      \includegraphics[width=\linewidth]{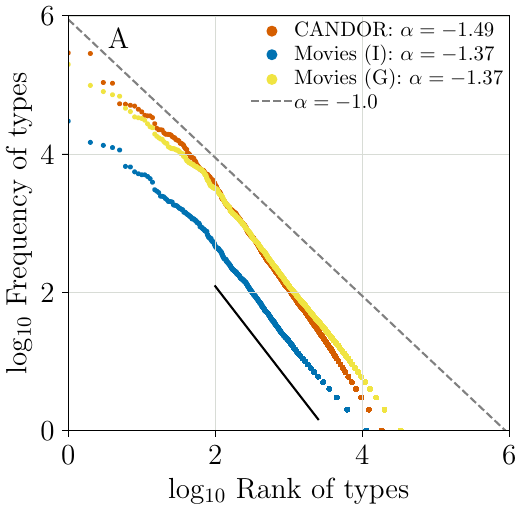}
    \end{minipage}
    \hspace{1em}
    \begin{minipage}[c]{0.38\textwidth}
    \vspace*{.4em}
      \includegraphics[width=\linewidth]{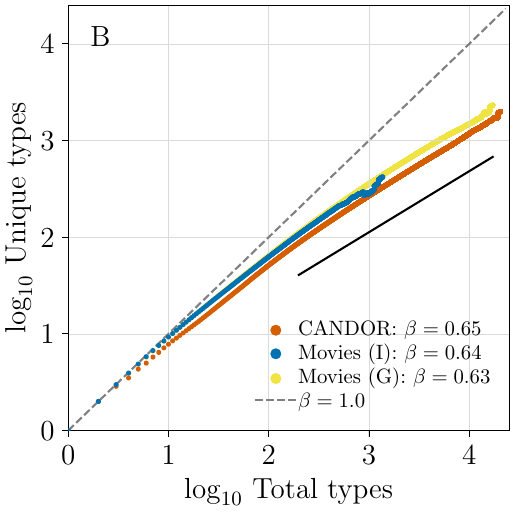}
    \end{minipage}
    \caption{Zipf's law (A) and \heapslaw\ (B) for the corpora. The grey dashed line indicates (coefficient $=1$), and the solid black line indicates the scaling regime restriction on the (A) horizontal axis or (B) vertical axis. \standrderrorstatement.}
    \label{fig:all_regime_scaling}
\end{figure*}

Looking at proportion of POS usage by corpus, `other', noun, and verb are respectively the largest categories for total usage while noun and verb are the largest categories for unique usage.
Unique noun proportions are: \moviesgroupshort\ ($0.39$), \candor\ ($0.32$), and \moviesindivshort\ ($0.29$).
Further differences are more apparent in function words and interjections: \moviesindivshort\ has the most unique function word usage ($0.09$ versus \candor\ $=0.06$ and \moviesgroupshort\ $=0.04$; see Table~\ref{table:interjections_freqs} for top occurring interjections).
\candor\ shows the most total interjection usage, but this proportion shrinks in unique usage ($0.11 \rightarrow 0.05$); \moviesindivshort\ and \moviesgroupshort\ interjection usage remains at $0.03$ for total and unique proportions.
When aggregating our POS categories of `other', function words, and interjections,
\candor\ accounts for $0.58$ of total words
(compared to $0.48$ in both \moviesindivshort\ and \moviesgroupshort); \candor\ differs from Ref.~\cite{chacoma_tags_2020}'s `other' proportion ($0.50$). This difference results from \candor's noun proportion ($0.22$) being lower than Ref.~\cite{chacoma_tags_2020} ($0.31$). Verbs are comparable across our sources to Ref.~\cite{chacoma_tags_2020}.

\subsection{Temporal analyses}
\label{sec:temporal}

We analyze conversations' narrative thirds (tertiles). For tertiles' interarrival distributions, all corpora's gaps are highly skewed short but with long tails (Figure~\ref{fig:narrative_tertile_intervals}). \candor\ and \moviesgroupshort\ resemble each other in this regard, though \candor's tail spans slightly more magnitudes in the beginning and middle of conversations. We sum the tails' word appearances as POS proportions by corpus (see Table~\ref{table:interarrival_rare_words_pos}). \candor's and \moviesgroupshort's highest tail gaps are nouns and \moviesindivshort's are `other'. We see examples of punctuation and pronouns as some of the shortest intervals.

We split interarrival distributions by POS (Figure~\ref{fig:pos_arrival_boxplots}). POS interval differences become more apparent when comparing total types to unique types. Noun, verb, and `other' POS total gaps are much shorter than function word and interjection usage (but with longer tails).

\candor\ has the most regular interjection usage, demonstrated by shorter gaps. Compared to other POS, interjections and function words show much longer wait times for new unique types. 
Unique types' interquartile range reflects extremely long intervals before new types appearing, capturing \heapslaw\ behavior.

We calculate burstiness (B) and memory (M) along with multiple other comparison groups as we have not seen many textual results in this dual-metric format, excepting Ref.~\cite{goh2008burstiness_memory} (with figures in Appendix~\ref{sec:appendix.variation_in_novelty}). In Figure~\ref{fig:burstiness_memory_combined}, we include B and M for: corpora, POS, literary works as a comparison (Frankenstein, Pride \& Prejudice, and Moby Dick), and shuffled versions of each body denoted with $\ast$.
We find differences for these literary texts compared to the 2 books reported in Ref.~\cite{goh2008burstiness_memory} which have burstiness and memory of effectively $0$.
Our literary comparisons cluster around $\text{B}=-0.09$, $\text{M}=0.35$.
\candor\ and \moviesgroupshort\ are close in burstiness to this literary cluster but lower in memory. \moviesindivshort\ (and its POS) shows the least memory (close to $0$) but the most regularity (negative burstiness). Parts of speech differ more starkly from \candor\ and \moviesgroupshort\ averages. Their interjections and function words are close to $\text{B}=0$ and $\text{M}=0$. The rest of the POS cluster around $\text{M}=0$ but trail into the quadrant of low memory and low burstiness, suggesting more alternation than random between interarrival times and more regular behavior.
Notably, all the \moviesindivshort\ POS are in the lower left quadrant.

\subsection{Regression analyses}
\label{sec:regressions}

We observe sublinear scaling and \heapslaw\ similarly to prior works, e.g., \cite{chacoma_tags_2020,linders2023zipf_spoken_dialog}.
Scaling exponents for individual conversations are normally distributed (Figure~\ref{fig:heaps_distrib}), consistent with multi-participant internet forums~\cite{kubo_2011} and with scaling behavior for spoken settings~\cite{abe_2021,bian2016scaling_like_lin}.
The three corpora show \candor:~$\heapsexponent = 0.65 \pm 0.0001$, \moviesindivshort:~$\heapsexponent = 0.64 \pm 0.003$, and \moviesgroupshort:~$\heapsexponent = 0.63 \pm 0.0003$ (see Figure~\ref{fig:all_regime_scaling} for Heaps' and corresponding rank regressions).

\begin{figure*}
    \centering
    \includegraphics[width=.73\linewidth]{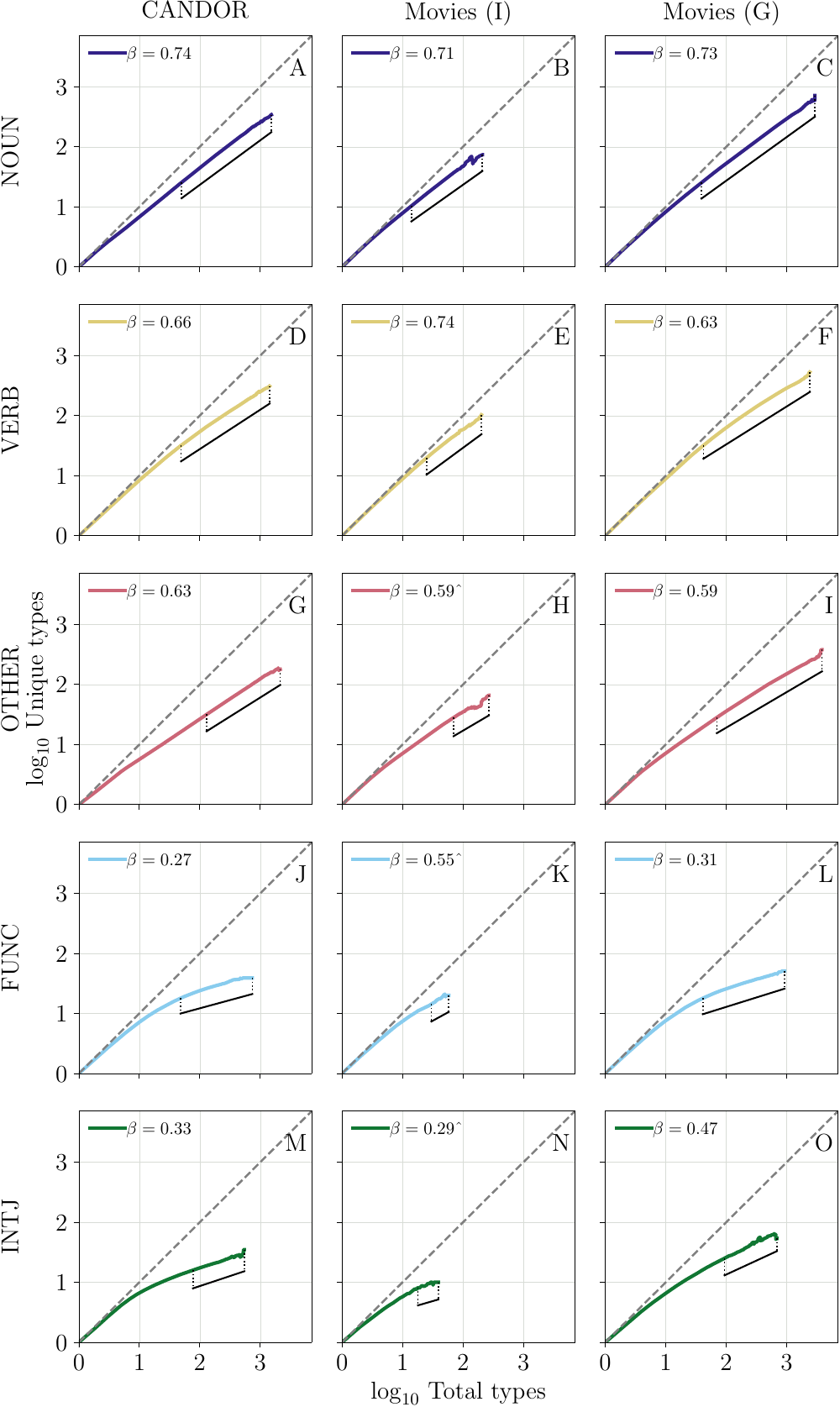}
    \caption{Part of speech linear regressions for the corpora. Below the dashed grey line ($\heapsexponent=1$) is sublinear scaling. The solid black line indicates the scaling regime restriction on the vertical axis. \standrderrorstatement.}
    \label{fig:pos_regressions}
\end{figure*}

\begin{figure*}
    \centering
    \includegraphics[width=.73\linewidth]{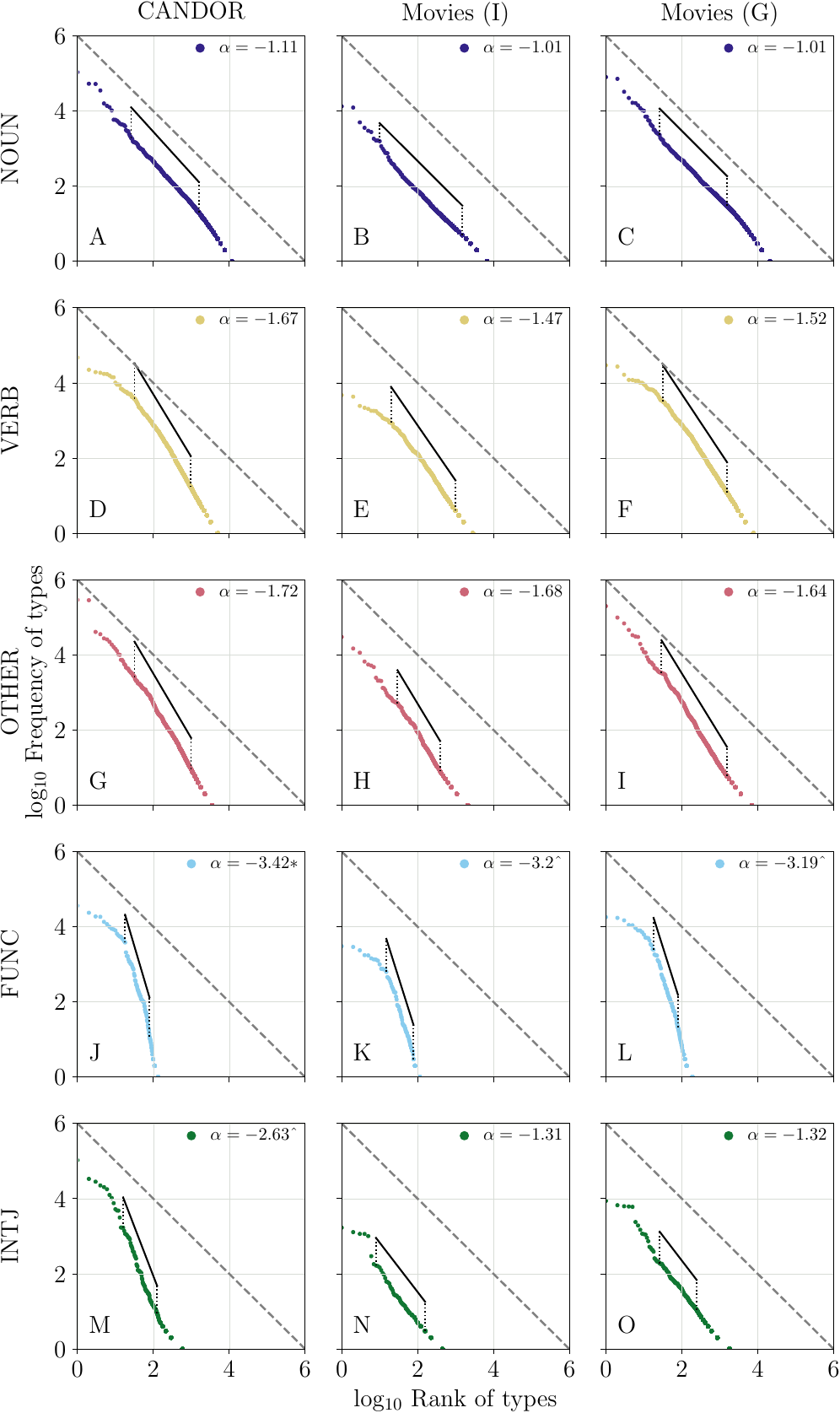}
    \caption{Part of speech rank regressions for the corpora. The dashed grey line indicates $\alpha=-1.0$. The solid black line indicates the scaling regime restriction on the horizontal axis. \standrderrorstatement.}
    \label{fig:pos_zipf_regressions}
\end{figure*}

Following the same process, we calculate linear regressions per POS per corpus on restricted scaling regimes (Figs.~\ref{fig:pos_regressions} and~\ref{fig:pos_zipf_regressions}).
\movies\ shows the highest standard errors of the slopes given their shorter and more variable length conversations. We note much higher standard error of the slope for \moviesindivshort\ when we keep punctuation in the analysis, which may be due to scaling regime changes from pre-punctuation (i.e., sensitivity to regime definition).
Nouns' coefficients are high across corpora (\candor\ being the highest) and higher than the corpora's overall $\heapsexponent$ values. Every other POS shows stark differences by corpora.
Verb and function word scaling are highest for \moviesindivshort.
\candor\ shows the highest coefficient for `other' POS.
Interjections are highest in \moviesgroupshort. 

Last, we regress on low and high interjection categories. The group proportions for low and high interjections are:
\candor\ $=[0.51,0.49]$, \moviesindivshort\ $=[0.60, 0.40]$, and \moviesgroupshort\ $=[0.50, 0.50]$. Split interjection analyses show little variation from their average corpora coefficients and by group. Reported as [low $\heapsexponent$ and high $\heapsexponent$]:
\candor~$= [0.68,0.67]$, \moviesindivshort~$= [0.75,0.72]$, and \moviesgroupshort~$= [0.68,0.65]$.

\section{Discussion}
\label{sec:papertag.discussion}


\subsection{Presence of \heapslaw\ in conversation}
Our work identifies \heapslaw\ scaling in two forms of conversational text: naturalistic conversation through video and fictional conversation via movies. We expected corpora level differences in scaling and found the datasets were similar at the corpus level; however, we do see differences in descriptive statistics.
That the trajectory of a movie plot (as communicated through conversations) has scaling in common with naturalistic conversations suggests a similar information efficiency mechanism. Of course, many of the same rules shape the language used in both contexts; for example, common words will be introduced early and repeated.

This scaling commonality at the corpus level is qualified by underlying differences in variance, which we suggest stems from lexical diversity. Initially, a plausible explanation for \moviesindivshort's higher variance seemed to be the difference in authorship between naturalistic video and movies. Movies' authorship often stems from collaboration of large writing teams with disparate artistic visions over extended periods of time. This collaborative authorship may be seen as similar to a multi-participant discussion. Research on multi-participant discussions suggests that having more participants could contribute to higher scaling behavior~\cite{kubo_2011,abe_2021}, but we do not see higher novelty production in movies' multi-authored scripts than in naturalistic conversation. Instead, we see statistical similarity between naturalistic video and conjoined movie scripts in total word counts, type-token ratios, and variance in \heapslaw\ exponents among their conversations.

However, single movie conversations' higher scaling variance and lexical diversity may serve to differentiate plot points to an audience. This purpose could partially explain this corpus' high variance while still following scaling behavior at its corpus level.
In particular, \moviesindivshort\ conversations may, by design, rely on a pairing of verbal and multimodal context to scaffold meaning~\cite{clark1996using_lang}: by pairing high lexical diversity with layers of multimodal content (e.g., cinematography, music), conversations differentiate plot points and imbue the intended story development into each utterance~\cite{zago2020film_discourse,martin2010pos_bnc_movies}.
This kind of scaffolding could feasibly help the audience make sense of holistic themes, such as coming of age or grief.
Meanwhile, naturalistic speech has different layers of content---gestural and prosodic cues---with spontaneous sense-making among participants.
This explanation is not necessarily incompatible with multiple authorship, as that in itself could contribute to lexical diversity as well, but future work would need to specifically select movies to compare authorship features.



\subsection{Part of speech differences by medium}
Having identified global corpus scaling behavior, we next focus on part of speech scaling and discuss lexical efficiency. We find open POS (nouns and verbs) to have higher scaling exponents than other POS categories---similarly to 
literary works~\cite{chacoma_tags_2020} and consistent with our expectation to see a difference between open and closed classes. Nouns exhibit relatively higher scaling behavior across our datasets than do other POS, with nouns being higher than average corpora exponents. Only verbs follow this higher-than-average pattern in \moviesindivshort.
This commonality to scaling in literary works~\cite{chacoma_tags_2020} suggests some universality in the way the English language informs listeners or audiences, with POS showing differences by medium.
Higher production of nouns (\candor, \moviesgroupshort) and verbs (\moviesindivshort)
may stem from length and structure of the message.
The former two corpora are longer and have potentially more information to convey;
in comparison, individual movie conversations are very short and may reference actions or events in order to be as informative as possible~\cite{martin2010pos_bnc_movies}.
Consider the proportions of unique nouns and verbs: these POS appear the most across corpora, with \moviesgroupshort\ showing higher unique nouns. Conjoined movie conversations likely rely on nouns to encode novel plot content over the course of films beyond single movie conversations and naturalistic conversations.
These speaking parts represent a complex phenomenon very different to spontaneous speech where the speech and world are co-created and intended to co-occur to progress the plot.
Nouns encode the arrival of new plot points and serve a pragmatic function in relation management: showing how speakers relate to each other socially and to their own stances~\cite{ruhlemann2007ConversationContext}, including speech acts~\cite{martin2009teaching_convo_film}, in a conversational context.

In \moviesindivshort, function word scaling is much higher than in the other two corpora, which we suggest helps to structure meaning for viewers.
Higher function word scaling may reflect a preference for greater clarity within shorter bouts of dialogue and
provide more complete context rather than relying on ellipsis, the omission of words that one can infer through context~\cite{merchant2018ellipsis_chapter}.
More complete context of function words can minimize cognitive load by working around an information locality constraint~\cite{grice1989wayofwords,futrell2024linguistic_info_efficiency}; for example characters reference earlier events in sufficient detail to further the more current plot point and achieve audience comprehension.
Notably, our exploration into burstiness and memory yields a similar result in \moviesindivshort, consistent with function words' more regular occurrence~\cite{altmann2009beyond_word_freq}.

For interjections, \moviesgroupshort\ shows the highest scaling, but \candor\ shows the highest total interjection use, confirming our expectation to see interjections more frequently in \candor. This finding suggests interjections serve a connective purpose in spontaneous versus pre-planned conversation.
Interjections' feedback and reinforcement role in conversation~\cite{cooney2024naturalturn,boudin2024multimodal_feedback_in_convo} is not functionally needed in prepared dialogue, which is more cohesively organized~\cite{martin2009teaching_convo_film,zago2020film_discourse}.
Further, the corpora rely on different interjections (Table~\ref{table:interjections_freqs}).
For instance, naturalistic speech has a higher volume of space-filling words (`like', `um') while fictional speech shows more introduction of concepts or contradiction (`well') and a general use of `yes' and `no'. It is possible that movies distribute unique interjections across functions other than space-filling, lending to higher scaling behavior.

Last, we end on how the affordances in each corpus help convey meaning, returning to lexical efficiency playing a role in shaping communication~\cite{zipf1949least_effort,fedzechkina2012restructure_lang_efficiency,jaeger2011efficiency_shape_lang_product,ferrer_i_cancho2003least_effort,xu2024wordreuse}.

Each corpus exhibits high use of a part of speech suggesting that novel word-forms are introduced more in that particular POS to encode information.
We suggest these higher-use parts of speech are less lexically efficient, offering high utility in communicating precisely in the respective modality. These POS are: nouns and verbs in \moviesgroupshort, function words in \moviesindivshort, and interjections in \candor.

An explanation for this lexical inefficiency could be the tendency to provide the most relevant information according to the delivery method;
under the maxim of quantity~\cite{grice1989wayofwords}, the speaker has to judge how to provide maximally useful content so the listener comprehends the intent.
A speaker may be relying on lexical competence strategies~\cite{marconi1997LexicalCompetence} to adapt to the affordances or circumstances under which they are communicating.
Given naturalistic (video) and entire movie scripts reliance on content words (nouns and verbs), speakers could be relying on referential competence to provide more pointers to the world and concepts. For instance, movies in their entirety have the benefits of uninterrupted time and planned multimodality to convey messages.
Single movie conversations' reliance on verbs also makes use of the referential lexical strategy but leans more heavily on inferential competence through function words; these conversations are short and have the challenge of connecting to both past and future points in the movie. These single conversations could be maximizing the space via function words' structure.

\subsection{Limitations and future work}
\label{sec:papertag.discussion.limitations}

We discuss limitations in our work and areas to build on, with applications in NLP, film analysis, and conversation research.

The parts of speech classification method we apply relies on linguistic dependency trees (treebanks). The stanza model for POS tagging errs in the direction of context from trees rather than classification based on surface-form (e.g., lexicon). Researchers could compare stanza output to other POS processors such as NLTK, spaCy, or large language model tagging to see how the models contribute to understanding of forms of text conversation.

In all data projects, text cleaning procedures shape the data. For instance, NLP analyses often remove stopwords and punctuation as a rule, but any text cleaning should be thoughtful to the purpose of the research.
We run analyses with and without punctuation, ultimately using analyses with punctuation because punctuation can convey content from prosody~\cite{wolf2023prosody_text_redundancy} and can more completely express Zipf's law~\cite{stanisz2024nlp_complexity}.
Further, we clean the mediums to be more comparable under the assumption that grammatical artifacts would not be acted or spoken.
However, film analysis could transcribe dialogue as performed to capture all signal, including deviations from official scripts and naturalistic features injected by actors (pauses, backchannels, and interjections). One could include these linguistic features as a `type' to analyze scaling behavior.


In using the Movie-Dialogs corpus, movie conversations were skewed in time and toward short conversations. Over three decades, social media text length consistently decreased and communication norms changed~\cite{di_marco2024linguistic_simplification}, suggesting that research in fictional storytelling could study similar evolutions to ask whether movie conversations are globally short or how script construction has changed.
Movie title sampling would also be improved by adding longer movie conversations and applying title selection criteria (e.g., `top of' list per year such as grossing or rated).

We compare scripted to naturalistic conversation in video format, yet much communication occurs in other venues.
One could compare these mediums to offline spoken conversation, literary works, texting~\cite{azios2023textingconvos}, and online communication examples to better situate conversation data from truly naturalistic to non-naturalistic.
Though non-naturalistic, online discussion forums share important qualities with naturalistic conversations that are absent from scripted conversations. For example, Reddit has a conversational posting style with community-specific speech, where users can supplement text with symbolic communication like emojis or GIFs. Discussion forums such as Reddit are further susceptible to post edits based on platform feedback, making the language susceptible to co-construction at the platform level.
Comparing a broader group of mediums could inform our understanding of communication when dialogue is spontaneous versus pre-planned, having implications for how mediums' afforadances cognitively and behaviorally drive conversation.

We offer two directions related to \heapslaw\ and conversation research.
We interpret differences among scaling law coefficients (regimes) as a window to narrow our focus into explanatory mechanisms.
This method analyzes a subset of data using a visual method to identify the data range, which is sensitive to the selected range (choice of $x_{\text{min}})$. If the research goal is to evaluate goodness of fit to the power law distribution, a more expansive method may be desired~\cite{clauset2009powerlaw_regimes} that does not commit to an a priori data range.

Further study of interjections could bolster signal in conversation research, as we have tagged interjections only as a part of speech and find that analyzing their word-forms may miss informational context.
Interjections could be disambiguated into words versus interjectional phrases to study role dependence in their use---such as in the roles of continuers, expressing a reaction, or acknowledgment.
We find more regular interjection cadence in naturalistic video conversation than in scripted conversation through the use of pro-social filler words. Given that interjectional phrases are connected to local contexts~\cite{boudin2024multimodal_feedback_in_convo}, it would be interesting to pair interjection-form and timing to see how they demonstrate movement between conversational purposes (such as showing interest or building common ground).
\section{Concluding remarks}
\label{sec:papertag.conclusion}

We analyze \heapslaw, finding the statistical law's scaling appears in video-format naturalistic and fictional conversations and that the behavior differs by parts of speech.
These differences suggest that statistical scaling can reveal modality-specific patterns through parts of speech, showing how these types of conversation encode information under differing purposes and modality constraints.
Media-specific constraints shape these differences, so it is important to understand the afforadances that may shape how we communicate and connect across technological contexts.
The various media pathways for communication are inherently unique in structure and usage, shaping conversation within each modality, and potentially reinforcing thought and behaviors within and between communities.
We should ask how those affordances impact our need for social connection~\cite{baumeister1995needtobelong,benus2014social_entrainment,sun2020_wellbeing,wheatley2024} and more broadly support or detract from the social health of our communities~\cite{holt2017soc_connection_priority}.
Fictional conversations with their authored intentions, naturalistic video conversations, and other online conversational modalities offer ways to analyze these differences. Overall, conversation is rich with incredible context to study how we make sense of each other and coalesce around meaning under different technological constraints.

\section*{Acknowledgments}

We are grateful for discussion with Alejandro Ruiz, Sarah Phillips, Carter Ward, Fitz Keenan-Koch, Bradford Demarest, and Ben Cooley. 
Our work was supported in part by 
MassMutual, 
National Science Foundation Award \#2242829, a philanthropic gift from an anonymous source, and the Spanish Ministry of Science.

\bibliography{convo-info}


\clearpage
\appendix
\section*{Supplementary Information}

\setcounter{page}{1}
\renewcommand{\thepage}{S\arabic{page}}
\renewcommand{\thefigure}{S\arabic{figure}}
\renewcommand{\thetable}{S\arabic{table}}
\setcounter{figure}{0}
\setcounter{table}{0}

\renewcommand{\thesection}{S\arabic{section}}
\setcounter{section}{0}

\addcontentsline{toc}{chapter}{\hspace{1cm} Appendix}
\label{sec:papertag.appendix}

\section{Descriptive Statistics}
\label{sec:appendix.descrip}

The normal distribution of conversation scaling provides perhaps a normative finding for naturalistic conversations and could be studied further across naturalistic samples. These distributions are from conducting one regression per conversation assuming the conversation was long enough to be included; thus, \moviesindivshort's range is very large. Corpus ranges for scaling exponents are:
\candor: $[0.54,0.81]$,
\moviesindivshort: $[0,1.78]$, and
\moviesgroupshort: $[0.57, 0.97]$.

\begin{figure}[bp!]
    \centering
    \includegraphics[width=.82\columnwidth]{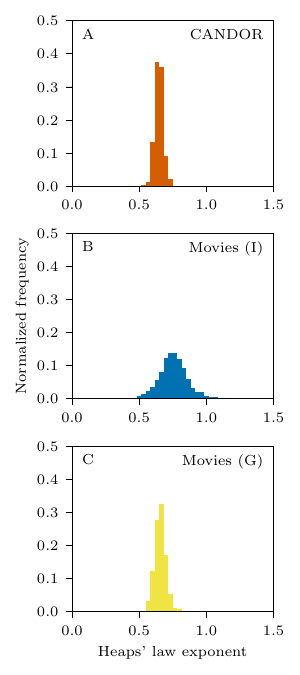}
    \caption{Distributions of scaling exponents by corpus. Corpus conversation $N$s are \convoNlist.}
    \label{fig:heaps_distrib}
\end{figure}

Plots in Figure~\ref{fig:all_scatter_aligned} show the variable nature of each corpus. We report additional statistics in Table~\ref{table:yet_more_descriptives} that qualify the spread of the data.

\begin{table}[]
    \centering
    \begin{tabular}{c|c|c|c|}
    \textbf{} & \textbf{\candor} & \textbf{\moviesindivshort} & \textbf{\moviesgroupshort} \\
    \hline
    \textbf{\shortstack{Utterances \\ ~}} & $36 / 182 / 574$ & $10 / 12 / 89$ & $6 / 473 / 1286$ \\

    \textbf{\shortstack{Words \\ ~}} & $2131 / 5256 / 16$k & $21 / 135 / 1157$ & $61 / 4833 / 14$k \\

    \textbf{\shortstack{Speaker \\ words}} & $519 / 4331 / 40$k & $5 / 108 / 2587$ & $1 / 135 / 5911$ \\

    \end{tabular}
    \caption{Raw descriptive statistics of utterances and words by corpus and words by speaker. Values are reported as: `min / median / max'.}
    \label{table:yet_more_descriptives}
\end{table}

\begin{figure*}
    \centering
    \includegraphics[width=.90\textwidth]{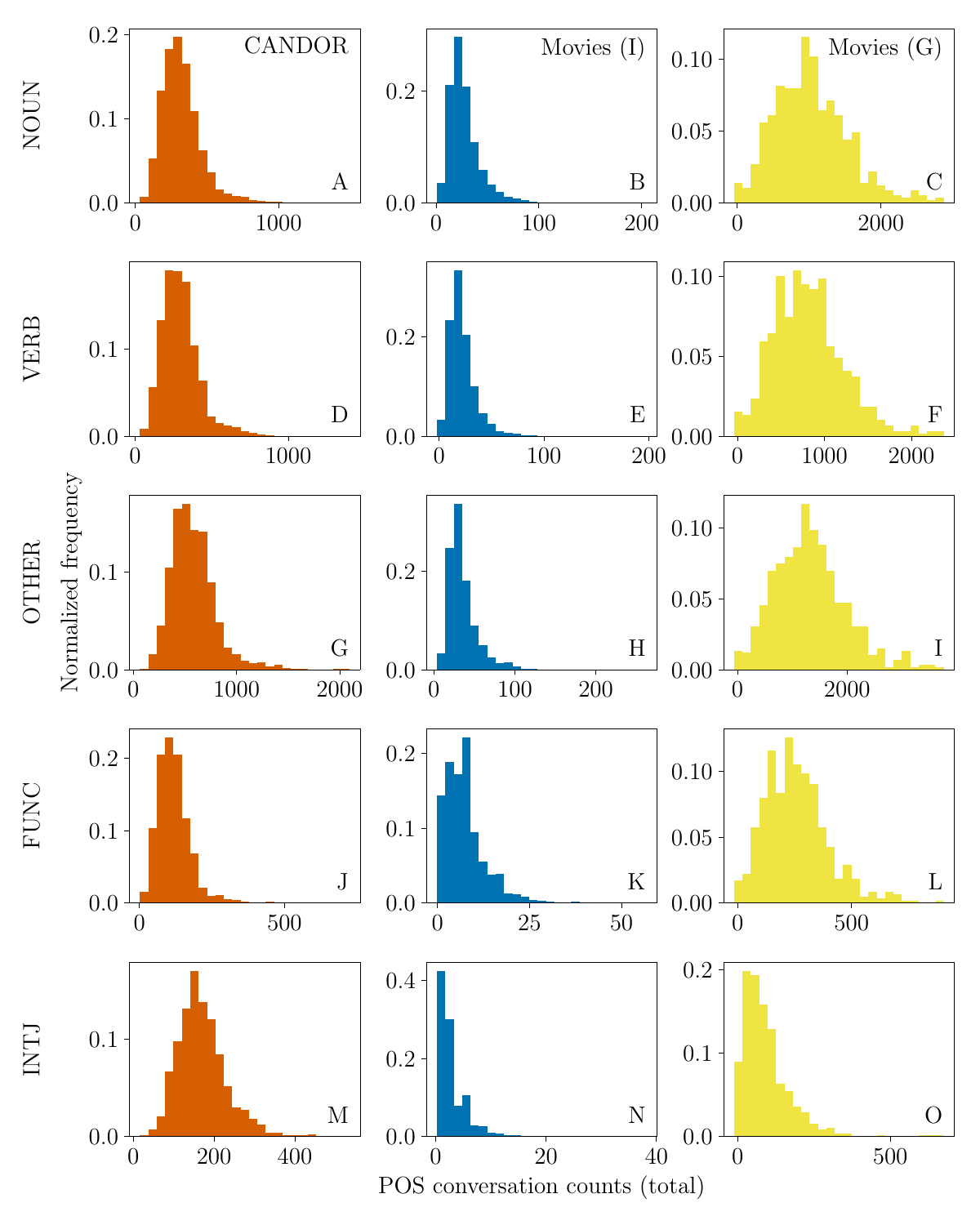}
    \caption{Part of speech conversation total count distributions by corpus. Each column is respectively \candor, \moviesindivshort, and \moviesgroupshort.}
    \label{fig:pos_total_distribs}
\end{figure*}

\begin{figure*}
    \centering
    \includegraphics[width=.90\textwidth]{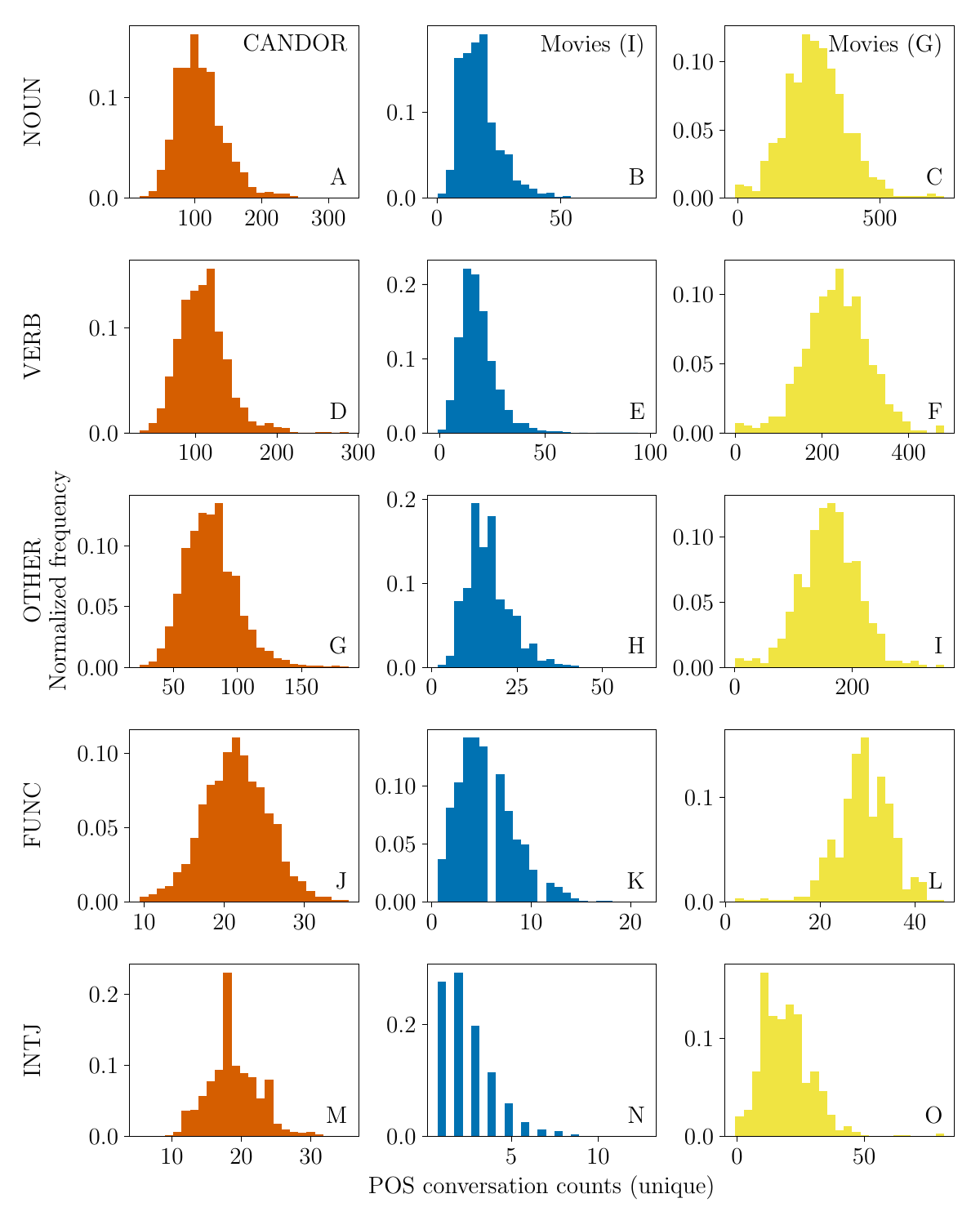}
    \caption{Part of speech conversation unique count distributions by corpus. Each column is respectively \candor, \moviesindivshort, and \moviesgroupshort.}
    \label{fig:pos_unique_distribs}
\end{figure*}

The next two plots show total (Fig.~\ref{fig:pos_total_distribs}) and unique (Fig.~\ref{fig:pos_unique_distribs}) conversation count distributions by corpus. Table~\ref{table:pos_descriptives} lists total (T) and unique (U) means and standard deviations, describing the raw data per POS per corpus. Figure~\ref{fig:pos_descriptives_normed} normalizes the proportion of each POS to give a better sense of their representation across conversation modes.

\bigskip
\bigskip
\begin{table}[h!]
    \newcolumntype{C}[1]{>{\centering\arraybackslash}p{#1}}
    \newcolumntype{R}[1]{>{\raggedleft\arraybackslash}p{#1}}

    \setlength{\tabcolsep}{0pt}
    \renewcommand{\arraystretch}{1.2}
    \begin{tabular}{R{1.9cm}|C{2.1cm}|C{2.1cm}|C{2.1cm}|}
    \multicolumn{1}{c|}{} & \textbf{\candor} & \textbf{\moviesindivshort} & \textbf{\moviesgroupshort} \\
    \hline

    \rowcolor{gray!15}
    \textbf{Noun } (T)~~ & 352.24 & 31.75 & 1082.45 \\
                        & (141.33) & (15.61) & (507.24) \\
    \rowcolor{cyan!10}
                    (U)~~ & 114.70 & 19.07 & 292.18 \\
                        & (36.59) & (8.24) & (108.74) \\

    \rowcolor{gray!15}
    \textbf{Verb } (T)~~ & 326.57 & 25.70 & 861.26 \\
                        & (129.92) & (13.22) & (408.65) \\
    \rowcolor{cyan!10}
                    (U)~~ & 114.46 & 19.71 & 242.45 \\
                        & (29.61) & (8.54) & (73.45) \\

    \rowcolor{gray!15}
    \textbf{Other } (T)~~ & 622.51 & 40.77 & 1399.96 \\
                        & (219.72) & (18.84) & (650.18) \\
    \rowcolor{cyan!10}
                    (U)~~ & 82.99 & 17.98 & 167.98 \\
                     & (21.15) & (6.50) & (50.28) \\

    \rowcolor{gray!15}
    \textbf{Func } (T)~~ & 132.51 & 8.60 & 276.52 \\
                    & (58.38) & (5.84) & (139.56) \\
    \rowcolor{cyan!10}
                    (U)~~ & 22.03 & 5.85 & 30.07 \\
                     & (3.94) & (2.96) & (5.90) \\

    \rowcolor{gray!15}
    \textbf{Intj } (T)~~ & 180.96 & 3.68 & 105.29 \\
                        & (61.94) & (2.99) & (80.23) \\
    \rowcolor{cyan!10}
                    (U)~~ & 19.38 & 2.61 & 21.01 \\
                     & (3.75) & (1.61) & (10.13) \\

    \end{tabular}
    \caption{Raw part of speech descriptive statistics by corpus for total words (T) and unique words (U). Values are $\mu$ on top with ($\sigma$) below its mean.}
    \label{table:pos_descriptives}
\end{table}

\clearpage

\subsection{Top interjections by corpus}

Within corpus, we calculate the most frequent interjections tagged using Stanza. After sorting on \candor, we use each corpus's interjection total to calculate percentages: \candor: $299,491$, \moviesindivshort: $9,812$, \moviesgroupshort: $61,913$. Any proportions less $0.01$ indicate so while zeros are indicated with a `-'. See Table~\ref{table:interjections_freqs} for the top 25 per dataset.
The top frequencies demonstrate that the part of speech tagging model is not without error (tagging some different forms of parallel speech as interjections), but it also shows some obvious differences in type of speech between corpora. Notably, the more space-filling words in English are tagged in naturalistic speech (`like', `uh', `um') while we see more tags in fictional speech suggesting the introduction of concepts or contradiction (`well').

\begin{table}[tp!]
    \centering
    \newcolumntype{C}[1]{>{\centering\arraybackslash}p{#1}}

    \setlength{\tabcolsep}{0pt}
    \renewcommand{\arraystretch}{1.2}

    \begin{tabular}{C{1.4cm}|C{2.1cm}|C{2.1cm}|C{2.1cm}|}
    \textbf{text} & \textbf{Candor} & \textbf{Movies (I)} & \textbf{Movies (G)} \\
    \hline
    yeah & 0.35 & 0.12 & 0.1 \\
    oh & 0.11 & 0.1 & 0.1 \\
    like & 0.1 & 0.01 & $<0.01$ \\
    uh & 0.07 & 0.01 & 0.02 \\
    mhm & 0.07 & - & - \\
    um & 0.06 & $<0.01$ & - \\
    ok & 0.04 & $<0.01$ & - \\
    no & 0.04 & 0.17 & 0.14 \\
    well & 0.03 & 0.13 & 0.11 \\
    ok. & 0.02 & - & - \\
    wow & 0.02 & $<0.01$ & $<0.01$ \\
    yes & 0.02 & 0.12 & 0.1 \\
    mm & 0.01 & - & - \\
    huh & $<0.01$ & 0.02 & 0.02 \\
    mhm. & $<0.01$ & - & - \\
    hm & $<0.01$ & - & - \\
    hello & $<0.01$ & 0.01 & 0.01 \\
    yes. & $<0.01$ & - & - \\
    hi & $<0.01$ & $<0.01$ & 0.01 \\
    mm. & $<0.01$ & - & - \\
    hey & $<0.01$ & 0.02 & 0.03 \\
    gosh & $<0.01$ & - & - \\
    man & $<0.01$ & $<0.01$ & $<0.01$ \\
    no. & $<0.01$ & - & - \\
    yep & $<0.01$ & - & - \\
    ah & - & $<0.01$ & $<0.01$ \\
    alright & - & $<0.01$ & $<0.01$ \\
    boy & - & $<0.01$ & $<0.01$ \\
    c'mon & - & $<0.01$ & $<0.01$ \\
    damn & - & - & $<0.01$ \\
    god & - & - & $<0.01$ \\
    mr & - & 0.01 & 0.02 \\
    okay & - & 0.03 & 0.04 \\
    please & - & 0.02 & 0.02 \\
    shit & - & $<0.01$ & $<0.01$ \\
    sorry & - & - & $<0.01$ \\
    uhhuh & - & $<0.01$ & $<0.01$ \\
    yah & - & $<0.01$ & - \\
    \end{tabular}
    \caption{Top $25$ interjection frequencies by corpus, calculated as the top appearing within each corpus and divided by its respective total interjection count.}
\label{table:interjections_freqs}
\end{table}

\clearpage
\section{Variation in Novelty}
\label{sec:appendix.variation_in_novelty}

\begin{figure}[!b]
    \centering
    \includegraphics[width=\linewidth]{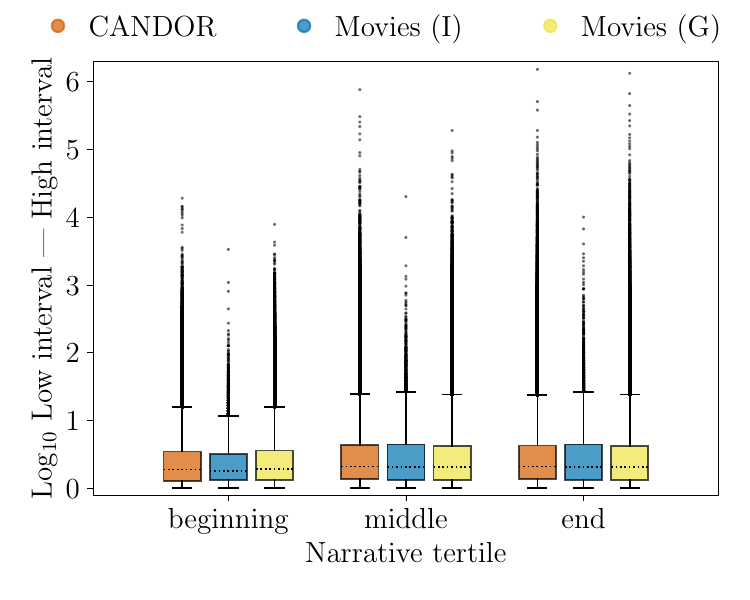}
    \caption{Interarrival times by corpus and narrative third in log$_{10}$ scale.}
    \label{fig:narrative_tertile_intervals}
\end{figure}

\begin{figure}[!t]
    \centering
    \includegraphics[width=\linewidth]{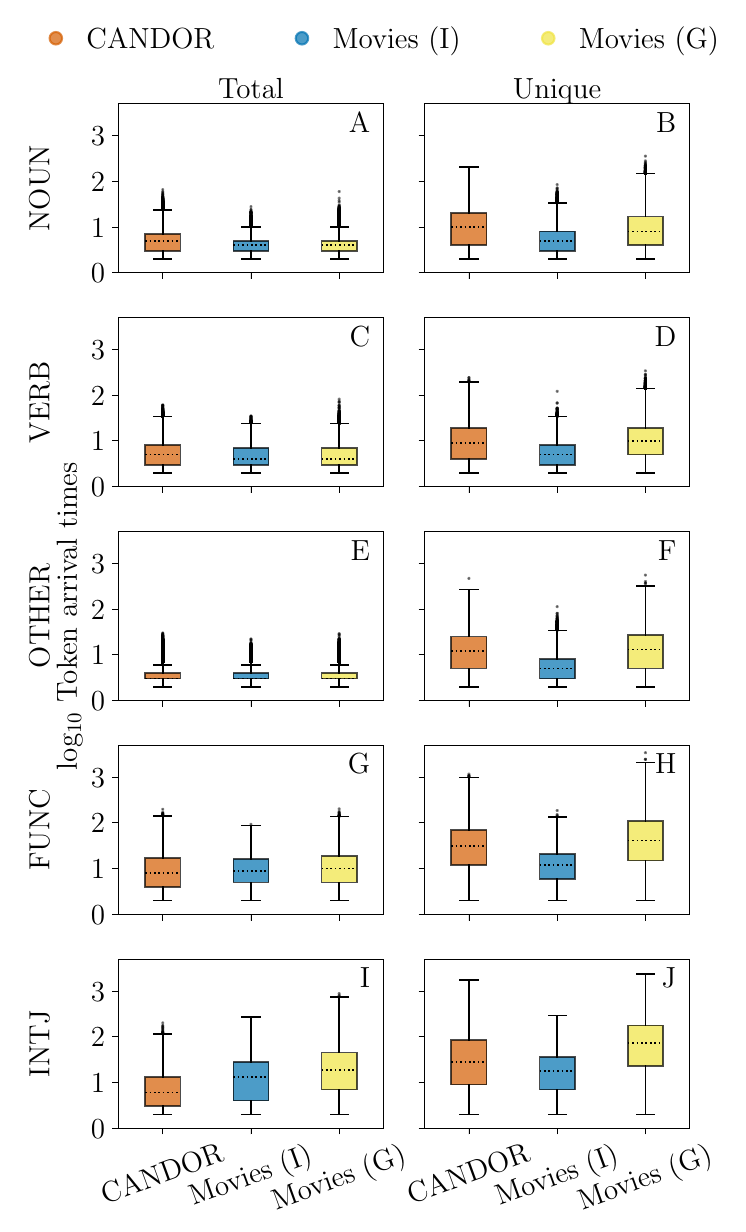}
    \caption{Interarrival times by corpus and part of speech in log$_{10}$ scale. The left column represents total tokens while the right column represents unique tokens.}
    \label{fig:pos_arrival_boxplots}
\end{figure}

Complementary measures can show the introduction of information similarly to \heapslaw. This exploration delves into the interarrival times of unique types and POS, measuring distributions as well as burstiness and memory. Interarrival times reflect the differences within index of occurrences of words and POS categories. For example, in the sequence:

[INTJ, FUNC, INTJ, FUNC, VERB, VERB, NOUN, FUNC, ...],
starting from the first function word occurrence (FUNC), interarrival times of 2 and 4 reflect differing POS until the next interjection appears (inclusive). Next we plot the interarrival gap distributions by corpus and section of text (Figure~\ref{fig:narrative_tertile_intervals}) as well as by POS (Figure~\ref{fig:pos_arrival_boxplots}).

We offer two methodological notes. First consider documents lengths: \moviesindivshort\ distributions are artificially shorter because the text lengths are much shorter. Document length also affects seeing extremely high arrival times of unique types because longer texts have more time for novel types to occur.
Second, calculating burstiness requires more than 2 items of the same kind because 2 items provide only 1 interval that results in mean and standard deviation of 0 and a burstiness value of -1, stacking extremely rare words in this region. We filter these rare word occurrences when calculating at the corpora level (on words), but this filtering inherently does not happen for POS because there are 5 categories that are filled every time and do not have rare types. Thus, the POS and corpora-level B and M \textit{averages} are slightly reflective of this difference while the corpora and literary cluster may be easier to compare.

\begin{table}[!b]
\centering
    \newcolumntype{P}[1]{>{\centering\arraybackslash}p{#1}}
    \newcolumntype{Q}[1]{>{\raggedleft\arraybackslash}p{#1}}

    \setlength{\tabcolsep}{0pt}
    \renewcommand{\arraystretch}{1.2}

    \begin{tabular}{Q{2cm}|P{1.8cm}|P{1.8cm}|P{1.8cm}|}
    \multicolumn{1}{c|}{} & \candor & \moviesindivshort & \moviesgroupshort \\
    \hline
    \textbf{Noun}~~~ & .31 & .35 & .51 \\
    \textbf{Verb}~~~ & .28 & .01 & .20 \\
    \textbf{Other}~~~ & .24 & .60 & .14 \\
    \textbf{Func}~~~ & .08 & .02 & .04 \\
    \textbf{Intj}~~~ & .08 & .01 & .11
    \end{tabular}
    \caption{Part of speech proportions for the highest interval words by corpus (tail-end of interarrival distributions). The log$_{10}$ cut off values were respectively [4.2, 2, 4.2] for \candor, \moviesindivshort, and \moviesgroupshort.}
    \label{table:interarrival_rare_words_pos}
\end{table}

\begin{figure*}
    \centering
    \begin{minipage}{\textwidth}
        \centering
        \includegraphics[width=0.99\textwidth]{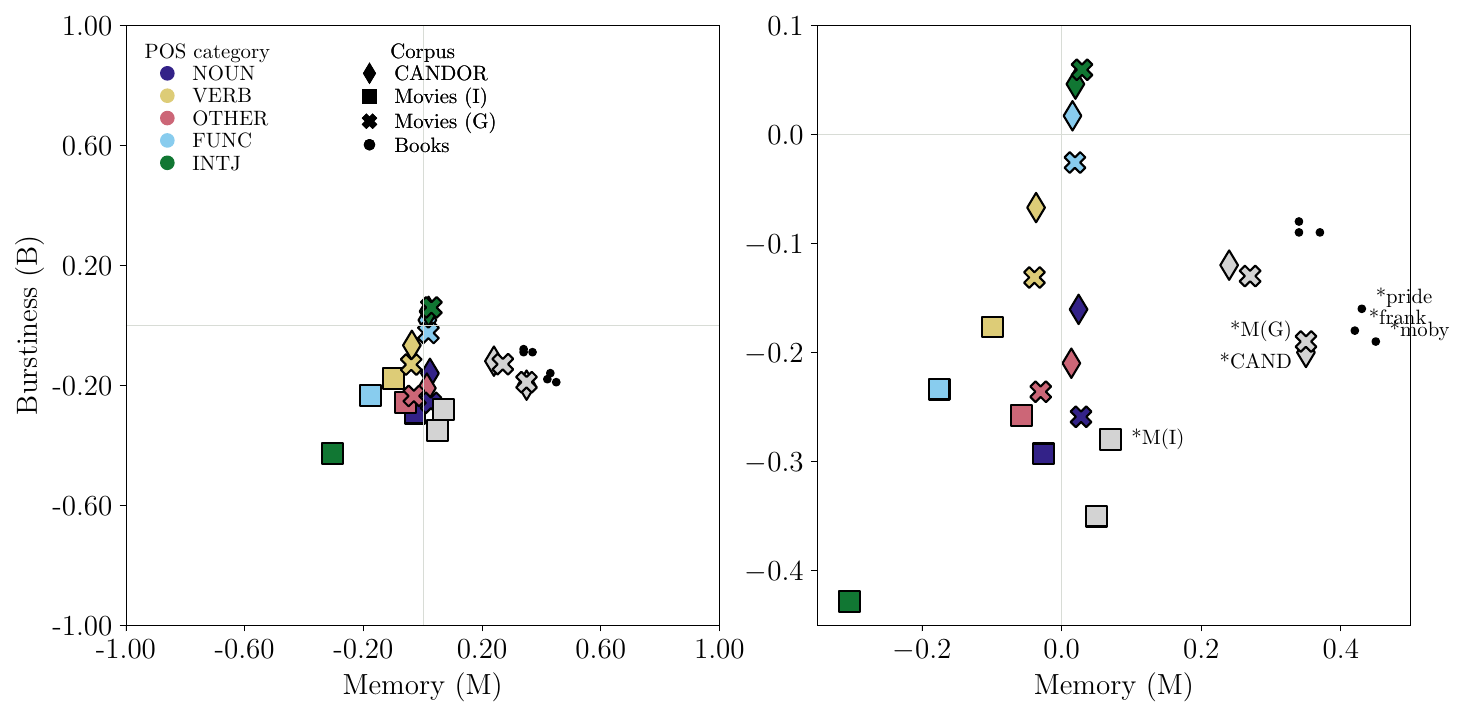}
        \vspace{1em} 

        \newcolumntype{P}[1]{>{\centering\arraybackslash}p{#1}}
        \newcolumntype{Q}[1]{>{\raggedleft\arraybackslash}p{#1}}

        \setlength{\tabcolsep}{0pt}
        \renewcommand{\arraystretch}{1.2}

        \begin{tabular}{Q{2cm}|P{1.9cm}|P{1.9cm}|P{1.9cm}|}
        \multicolumn{1}{c|}{} & \candor & \moviesindivshort & \moviesgroupshort \\
        \hline
        \textbf{Noun}~~~ & -.16 / .02 & -.29 / -.03 & -.26 / .03 \\
        \textbf{Verb}~~~ & -.07 / -.04 & -.18 / -.10 & -.13 / -.04 \\
        \textbf{Other}~~~ & -.21 / .01 & -.26 / -.06 & -.24 / -.03 \\
        \textbf{Func}~~~ & .02 / .02 & -.23 / -.18 & -.03 / .02 \\
        \textbf{Intj}~~~ & .05 / .02 & -.43 / -.30 & .06 / .03 \\
        \end{tabular}
        \vspace{1em}

        \caption{Burstiness and memory by corpora and parts of speech. Top: Burstiness and memory by corpora and parts of speech. Left is the full metric range from -1 to 1, and the right plot zooms into the data range. Asterisked labels are those corpora shuffled. Bottom: Burstiness and memory by corpora and part of speech with values reported as $\text{B}/\text{M}$. Please refer to the methodological note in this section about comparing corpora averages and POS.}
        \label{fig:burstiness_memory_combined}
    \end{minipage}
\end{figure*}

\clearpage
\section{Methodology Expansion}
\label{sec:appendix.methods_extra}

\subsection{Data cleaning}

We recode raw POS tags using the following schema~\cite{chacoma_tags_2020}: adverb $\rightarrow$ verb,
pronoun $\rightarrow$ noun,
proper noun $\rightarrow$ noun,
adposition $\rightarrow$ function words,
coordinating conjunction $\rightarrow$ function words,
subordinating conjunction $\rightarrow$ function words,
interjections remain interjections,
and all remaining tags $\rightarrow$ other.

\subsection{Regressions}

For corpus-level scaling regimes, we use log-space values of $2.0 < x \leq 3.4$.

With POS as rows (in order of Noun, Verb, Other, Func, Intj), and corpora as columns (in order of \candor, \moviesindivshort, and \moviesgroupshort), log-space start values are:
\[
\begin{bmatrix}
1.4 & 1.0 & 1.4 \\
1.5 & 1.3 & 1.5 \\
1.5 & 1.45 & 1.45 \\
1.25 & 1.15 & 1.25 \\
1.2 & 0.9 & 1.4
\end{bmatrix}
\]
end values are:
\[
\begin{bmatrix}
3.2 & 3.2 & 3.2 \\
3.0 & 3.0 & 3.2 \\
3.0 & 2.6 & 3.2 \\
1.9 & 1.9 & 1.9 \\
2.1 & 2.2 & 2.4
\end{bmatrix}
\]

We calculate average exponents over conversations for corpus-level findings. Initially our method involved one regression per conversation, but this approach disqualified numerous very short movie conversations from the corpus-level analysis. As a second approach, we initialize a matrix to be the maximum length of the longest conversation in a corpus, in which we store the number of unique types per conversation. Then, we calculate the average unique types at each step of $N$ up to the maximum based on the number of conversations having types at that `window', excluding zeros from the calculation (where conversations end). This approach works for \moviesindivshort's very short conversations, so for comparability of approach, we apply this method across all corpora. This methodological choice makes little difference in the larger corpora's coefficients: \\
corpus: $\heapsexponent_\text{former method} \rightarrow \heapsexponent_\text{final method}$.\\
\candor: $0.66 \rightarrow 0.65$.\\
\moviesindivshort: $0.76 \rightarrow 0.64$.\\
\moviesgroupshort: $0.67 \rightarrow 0.63$.

\begin{figure}[b!]
    \centering
    \includegraphics[width=.74\columnwidth]{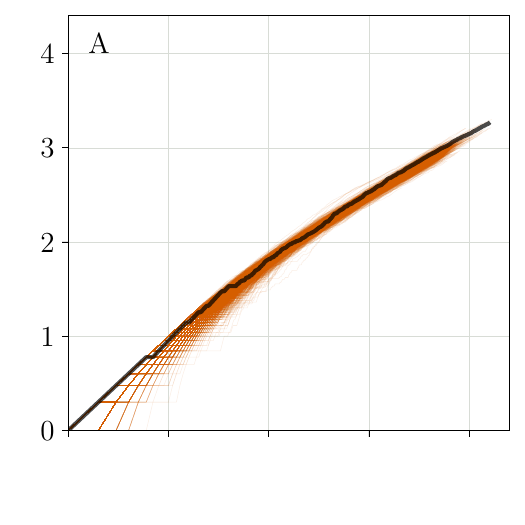}\\[-5.5ex]
    \includegraphics[width=.74\columnwidth]{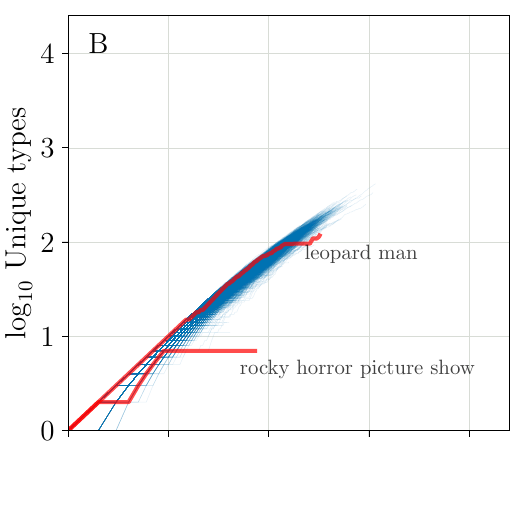}\\[-5.5ex]
    \includegraphics[width=.74\columnwidth]{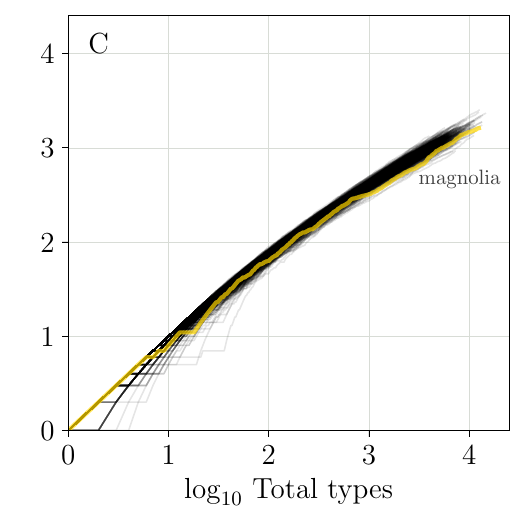}
    \caption{Conversation-level regressions by corpora. These plots show one regression per conversation, an alternative method we decided against in the current project. In A) CANDOR, B) Movies (I), and C) Movies (G), the stand-out line is an outlier conversation. }
\label{fig:regression_outliers_panel}
\end{figure}

\noindent Figure~\ref{fig:regression_outliers_panel} shows conversation-level regressions. We use a stand-out color per corpus to denote outliers where the conversation's innovation flattened for a minimum raw count of unique types in the conversation ($125$ for \candor\ and \moviesgroupshort\ and $60$ for \moviesindivshort). While steps or plateaus appear in the lower-left quadrant on these charts, these visual steps do not meet the threshold for outliers. These areas visually look like steps because raw data transformed to $log_{10}$ scale produce tiny decimal increments.

\end{document}